\documentclass[preprint]{elsarticle}

\usepackage{charter}
\usepackage{amsmath,amsfonts,amssymb,amsthm}

\theoremstyle{definition}

\usepackage[euler-digits,euler-hat-accent]{eulervm}
\usepackage{bm}
\usepackage{lineno,hyperref}
\usepackage{caption}
\usepackage{subcaption}
\usepackage{float}
\usepackage{tikz}
\modulolinenumbers[5]

\makeatletter
\def\ps@pprintTitle{%
	\let\@oddhead\@empty
	\let\@evenhead\@empty
	\def\@oddfoot{}%
	\let\@evenfoot\@oddfoot}
\makeatother

\begin{document}

\begin{frontmatter}

\title{Closed-form solutions for the inverse kinematics of serial robots using conformal geometric algebra}


\author[mymainaddress]{Isiah Zaplana\corref{mycorrespondingauthor}}
\cortext[mycorrespondingauthor]{Corresponding author}
\ead{isiah.zaplana@kuleuven.be}

\author[mysecondaryaddress]{Hugo Hadfield}
\ead{hh409@cam.ac.uk}

\author[mysecondaryaddress]{Joan Lasenby}
\ead{jl221@cam.ac.uk}

\address[mymainaddress]{Department of Mechanical Engineering, KU Leuven -- University of Leuven, Leuven, Belgium}
\address[mysecondaryaddress]{Department of Engineering, University of Cambridge, Cambridge, UK}

\begin{abstract}
This work addresses the inverse kinematics of serial robots using conformal geometric algebra. Classical approaches include either the use of homogeneous matrices, which entails high computational cost and execution time, or the development of particular geometric strategies that cannot be generalized to arbitrary serial robots. In this work, we present a compact, elegant and intuitive formulation of robot kinematics based on conformal geometric algebra that provides a suitable framework for the closed-form resolution of the inverse kinematic problem for manipulators with a spherical wrist. For serial robots of this kind, the inverse kinematics problem can be split in two subproblems: the position and orientation problems. The latter is solved by appropriately splitting the rotor that defines the target orientation in three simpler rotors, while the former is solved by developing a geometric strategy for each combination of prismatic and revolute joints that forms the position part of the robot. Finally, the inverse kinematics of 7 DoF redundant manipulators with a spherical wrist is solved by extending the geometric solutions obtained in the non-redundant case.
\end{abstract}


\end{frontmatter}

%\linenumbers

\section{Introduction}\label{Intro}
A serial robot is an open kinematic chain made up of rigid bodies, called \textit{links}, connected by kinematic pairs, called \textit{joints}, that provide relative motion between consecutive links. The point at the end of the last link is known as the \textit{end-effector}. Only two types of joints are considered throughout this work: \textit{revolute} (\textit{prismatic}) joints, that perform a rotational (translational) motion around (along) a given axis.

\begin{figure*}[t!]
	\centering
	\includegraphics[width=5cm] {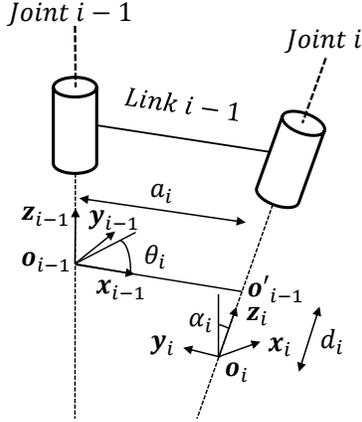}
	\caption{The four D-H parameters: the length of link $i$ ($a_{i}$), the angle between the joint axes $\bm{z}_{i-1}$ and  $\bm{z}_{i}$ ($\alpha_{i}$), the distance between $\bm{o}_{i-1}$ and  $\bm{o}_{i}$ along $\bm{z}_{i}$ ($d_{i}$) and the angle between $\bm{x}_{i-1}$ and  $\bm{x}_{i}$ ($\theta_{i}$). Clearly, if the $(i-1)$-th and $i$-th joint axes intersect (are parallel), then $a_i = 0$ ($\alpha_i = 0$) and $d_i$ ($a_i$) is the length of the $i$-th link.}
	\label{DH}
\end{figure*}

The end-effector position and orientation (also known as the \textit{pose}) is expressed as a differentiable function $f:\mathcal{C}\to X$, where $\mathcal{C}$ denotes the space of joint poses, known as the {\em configuration space} of the robot, while $X$ denotes the space of all positions and orientations of the end-effector with respect to a reference frame, known as the {\em operational space} of the robot. We say that the \textit{workspace} of the robot, denoted by $\mathcal{W}$, is the volume of the operational space $X$ that is reachable by the robot's end-effector. We attach a frame to each joint of the robot so that it describes the relative position and orientation of the joint. The relations between consecutive joint frames can be expressed by homogeneous matrices based on the Denavit-Hartenberg convention \cite{DenavitHartenberg65}. This convention consists of the use of four parameters, the \textit{D-H parameters} (figure \ref{DH}): one acting as a {\em joint variable} -- angle $\theta_{i}$ or displacement $d_{i}$, depending on the nature of the joint -- and the other three acting as constants -- length $a_{i}$, angle $\alpha_{i}$ and $d_{i}$ or $\theta_{i}$ depending on which one describes the joint variable \cite{Siciliano08a,SpongHutchinsonVidyasagar06}. In general, if the nature of the joint $i$ is not specified, its joint variable is denoted by $q_{i}$. The vector of all joint variables is denoted by $\bm{q}= (q_{1},\dots,q_{n})$ and is known as the {\em configuration}. Therefore, the configuration space $\mathcal{C}$ is the space of joint configurations. A robot is said to have \textit{$n$ degrees of freedom} (DoF) if its configuration can be minimally specified by $n$ variables. For a serial robot, the number and nature of the joints determine the number of DoF. For the task of positioning and orientating its end-effector in the space, the manipulators with more than 6 DoF are called \textit{redundant}, while the rest are \textit{non-redundant}.

Therefore, we can associate a homogeneous matrix $T_{i}^{i-1}$ with each joint frame $\{i\}$ such that it relates such frame to the preceding one, i.e.,  joint frame $\{i-1\}$. The first joint frame is related to the fixed reference frame. For each configuration $\bm{q}\in\mathcal{C}$, $f(\bm{q})$ can be represented using these homogeneous matrices as follows:
\begin{equation}\label{RelationMatrices}
T_{n}^{0} = T_{1}^{0}\cdot T_{2}^{1}\cdot\empty\dots\empty\cdot T_{n}^{n-1}
\end{equation}
with
\begin{equation}\label{frepresentation}
T_{n}^{0} = \begin{pmatrix}
R & \bm{p}\\
0 & 1
\end{pmatrix}
\end{equation}
where $R$ denotes the rotation matrix describing the end-effector orientation with respect to the fixed reference frame and $\bm{p}$ is the vector that describes the end-effector position with respect to that reference frame \cite{DenavitHartenberg65,Siciliano08a,SpongHutchinsonVidyasagar06}. An important class of serial robots are those with a spherical wrist. For these robots, the axes of their last three joints intersect at a common point, known as the \textit{wrist center point}, or are parallel (the intersection point and, hence, the wrist center point, is the point at the infinity). In this context, the end-effector position, $\bm{p}$, can be moved to the wrist center point $\bm{p}_w$ by a fixed translation and, thus, the last three joints would only contribute to the orientation of the end-effector.

This paper is focused on one of the most important problems in robot kinematics, namely the \textit{inverse kinematic problem}. It consists of recovering the joint variables, i.e., the configuration, given a target end-effector pose. This configuration may not be unique, since non-redundant manipulators have up to sixteen different configurations for the same end-effector pose \cite{Pieper68}, while for redundant manipulators this number is unbounded \cite{Siciliano08a,SpongHutchinsonVidyasagar06}. The opposite problem is known as the \textit{forward kinematic problem} and consists of obtaining the pose of the end-effector given the value of the joint variables.

The inverse kinematic problem plays a major role in robotics, specially in robot kinematics, motion planning and control theory. There are different methods used to solve it and they are usually categorized in two groups:
\begin{itemize}
	\item[a)] Closed-form methods: all the solutions are expressed in terms of the entries of $\empty^0T_n$. These methods strongly depend on the geometry of the manipulator and, therefore, are not sufficiently general. However, it is clear that they have advantages over the numerical methods such as, for instance, lower computational cost and less execution time. Additionally, they give all the solutions for a given end-effector's pose. Some closed-form methods include strategies based on machine learning \cite{Jiokoukouabon2020}, matrix manipulations \cite{Pieper68,Paul81}, the definition of the arm angle parameter \cite{JungYooKooSongWon11,YuJinLiu12} and different geometric methods \cite{HuangJiang13,LiuWangSunChangMaGeGao15,WeiJianHeWang14}.
	\item[b)] Numerical methods: a good approximation $\bm{\tilde{q}}$ of one of the solutions is derived iteratively. Although these methods usually work for any manipulator, they suffer from several drawbacks such as, for example, high computational cost and execution time, existence of local minima and numerical errors. Moreover, only one of the sixteen (infinite) possible solutions is obtained for non-redundant (redundant) manipulators. The most common numerical approaches are the Jacobian-based methods \cite{AristidouLasenby2018,Buss09,DeLucaOriolo91,LauWai02,WangSugaIwataSugano12,Wampler86}.
\end{itemize}
Among all the methods presented in this section, the closed-form methods are the most suitable for serial robots since they allow us to obtain the set of all solutions with a small computational cost. This paper proposes a formulation of the inverse kinematic problem based on conformal geometric algebra. This formulation allows us to avoid the use of matrices and, due to its inherent geometric nature, provides a geometrical description of the problem with which the inverse kinematics can be solved easily. These properties are exploited in this work to solve the inverse kinematics of serial robots with 6 DoF and spherical wrist. For these manipulators, this problem can be decoupled into two subproblems: the position problem and the orientation problem. For the former, a classification of the different combinations of prismatic and revolute joints that describe the position part of the manipulator is made and a geometric strategy is developed to solve each one of them, while, for the latter, we split the rotor that defines the target orientation into three simpler rotors, each one of them depending on a unique joint variable. Finally, the inverse kinematics of redundant serial robots with 7 DoF and spherical wrist is solved by extending the geometric solutions obtained in the non-redundant case.

The rest of the paper is organized as follows: Section 2 reviews the related work and presents a basic introduction to geometric algebra. In Section 3, a description of the forward kinematics of an arbitrary serial robot based on conformal geometric algebra is given. The formulation of the inverse kinematic problem and its solution are developed in Sections 4 and 5. In Section 6, the inverse kinematics of 7 DoF redundant serial robots is solved by extending the strategies introduced in Section 6. Finally, Section 7 presents the conclusions.  

\section{Related work and mathematical preliminaries}

\subsection{Related work}\label{RW}
As mentioned in the introduction, some closed-form methods are usually difficult to formulate and solve for non-redundant robots even if they have a spherical wrist. For instance, the Pieper method \cite{Pieper68} allows us to obtain the solutions of the inverse kinematics as the solutions of a set of polynomials and, for a general serial robot with a spherical wrist, at least one of these polynomials is of degree four, which is, in general, difficult to solve.

On the other hand, the Paul method \cite{Paul81} consists of the manipulation of the homogeneous matrices of the relation (\ref{RelationMatrices}) so the following family of matrix equations is obtained:
\begin{equation}\label{PaulCh4}
\biggl(T_{i-1}^{i-2}\biggr)^{-1}\cdots\biggl(T_{1}^{0}\biggr)^{-1}\cdot T_{n}^{0} = T_{i}^{i-1}\cdots T_{n}^{n-1}\quad\text{ for } i=2,\dots,n.
\end{equation}
The objective is to isolate those non-linear equations that contain just one joint variable. However, even if we find those equations, solving them analytically is, in general, difficult. Similarly, the formulation and implementation of geometric methods is also difficult. If, for instance, the circle intersection of a plane with a sphere is considered, then a system of two equations (one of them non-linear) must be solved. In this context, geometric algebra turns to be very useful. As stated in the introduction, geometric algebra and, in particular, the conformal model of three-dimensional Euclidean space, provides a framework for modeling the kinematics and dynamics of rigid bodies that avoids the use of matrices. In addition, the geometric entities such as points, lines, planes or spheres can be represented with elements of the algebra.

Regarding the contributions using this framework, in \cite{Bayro-CorrochanoKahler01,Selig01} analogous versions to the Paul and Pieper methods are presented. In both works, an easy and compact formulation of the problem is presented using the language provided by geometric algebra. However, they have the same drawbacks as the original methods. The approach presented in \cite{AristidouLasenby11} develops a fast and heuristic method that solves the inverse kinematics of arbitrary redundant serial robots. Nevertheless, the orientation problem is not treated and, since it is a numerical method, only one solution is obtained. Similarly, the approach presented in \cite{HildenbrandZamoraCorrochano08} only solves the position problem for an anthropomorphic manipulator.

Other approaches are focused on the development of geometric strategies based on conformal geometric algebra for particular robots. For example, in \cite{HrdinaNavratVasik16b}, a kinematic model of a planar redundant manipulator is developed, while in \cite{KimJeongPark15a} the inverse kinematics and the singularity problem are formulated and solved for a class of parallel robots. A 5 DoF serial robot is considered in \cite{ZamoraCorrochano04}, where the geometric strategy followed is the same as in \cite{HildenbrandZamoraCorrochano08}.  There, different geometric objects are defined. These objects, as well as the relations between them, are described using conformal geometric algebra. The angles between such geometric entities correspond to the unknown joint variables. Finally, the approach developed in \cite{Campos-MaciasCarbajal-EspinosaLoukianovCorrochano17} solves the inverse kinematics of a 6 DoF humanoid leg. In \cite{TordalHovlandTyapin16}, the authors solve geometrically the position problem of the 6 DoF Comau Smart-5 NJ-110, while in \cite{KleppeEgeland16}, particular solutions of the inverse kinematics of the UR5 and Agilus KR6 R900 are given. All of these works are based on the strategies introduced in \cite{HildenbrandZamoraCorrochano08,ZamoraCorrochano04}.

In this work, an approach to solve the inverse kinematics of arbitrary 6 and 7 DoF serial robots with a spherical wrist is presented. This approach is not focused on a particular robot but applies to an entire class of manipulators. Although not every single case is covered, the majority and most representative ones are treated in this work. This allows to illustrate the power of conformal geometric algebra when applied to the inverse kinematics of serial robots with spherical wrist as well as to provide a set of easy-to-implement geometric strategies which, in  turn, is the main goal of the present work \color{black}In particular, the position problem is solved geometrically by extending the particular contributions of \cite{HildenbrandZamoraCorrochano08,TordalHovlandTyapin16,LavorXamboZaplana18,Hugo2020,Zaplana20}, while a novel method is developed for solving the orientation problem. This method involves splitting the rotor that defines the target orientation into three rotors, so that each one depends on just one joint variable. The main advantage of this strategy lies in the fact that this approach is still a closed-form method, i.e., all the solutions are obtained as analytical expressions in terms of the end-effector pose. In addition, we are dealing with arbitrary serial robots of 6 and 7 DoF, not just a particular robotic geometry. Finally, due to the simplicity of the problem formulation and the geometric nature of conformal geometric algebra, such expressions are obtained easily than with other closed-form approaches (such as, for instance, the ones reviewed here).

\subsection{Mathematical preliminaries}\label{Preliminaries}
Geometric algebra provides strong advantages with respect to more common approaches such as, for instance, that geometric objects and Euclidean transformations live in the same algebra, that the expressions from the algebra are coordinate free and that it is well adapted to deal with more general fields in science and engineering. In particular, it provides a compact and neat formulation of robot kinematics and its main problems. Throughout this section, a brief overview of both geometric algebra and conformal geometric algebra is presented. A more detailed treatment of the subject can be found in \cite{DoranLasenby03,Dorst-Fontijne-Mann-2007,LasenbyLasenbyWareham04}.

\subsubsection{Geometric algebra}\label{GA}
Let us consider the real vector space $\mathbb{R}^{n}$ with orthonormal basis $\{e_{1},\dots,e_{n}\}$. The geometric algebra of $\mathbb{R}^{n}$, denoted by $\mathcal{G}_{n}$, is a vector space where the operations defined in $\mathbb{R}^{n}$, i.e., the addition and multiplication by scalars, are extended naturally. An additional operation, the {\em geometric product}, is defined on the algebra $\mathcal{G}_{n}$ and, acting on vectors is defined as: 
\begin{equation}\label{geometricproduct}
\bm{v}_{1}\bm{v}_{2} = \bm{v}_{1}\cdot \bm{v}_{2} + \bm{v}_{1}\wedge \bm{v}_{2},\,\text{ for } \bm{v}_{1},\bm{v}_{2}\in\mathbb{R}^{n},
\end{equation}
where $\cdot$ denotes the inner or dot product and $\wedge$ denotes the outer product. 

The outer product of two vectors $\bm{v}_{1},\bm{v}_{2}$ is a new element of $\mathcal{G}_{n}$, which is termed a \textit{bivector}, is said to have {\em grade} two and is denoted by $\bm{v}_{1}\wedge \bm{v}_{2}$. By extension, the outer product of a bivector with a vector is known as a \textit{trivector} and is denoted by $(\bm{v}_{1}\wedge \bm{v}_{2})\wedge \bm{v}_{3}$. Clearly, trivectors have grade three. This can be generalized to an arbitrary dimension. Thus,
\begin{equation}\label{rblade}
(\bm{v}_{1}\wedge \bm{v}_{2} \wedge \dots \wedge \bm{v}_{r-1})\wedge \bm{v}_{r}
\end{equation}
denotes an \textit{$r$-blade}, an element of $\mathcal{G}_{n}$ with grade $r$. 

A bivector $\bm{v}_{1}\wedge \bm{v}_{2}$ can be interpreted as the oriented area defined by the vectors $\bm{v}_{1}$ and $\bm{v}_{2}$. Thus, $\bm{v}_{2}\wedge \bm{v}_{1}$ has opposite orientation and, from that, the anticommutativity of the outer product can be deduced. Analogously, a trivector is interpreted as the oriented volume defined by its three composing vectors. Since the volume generated by $(\bm{v}_{1}\wedge \bm{v}_{2})\wedge \bm{v}_{3}$ is the same as the volume generated by $\bm{v}_{1}\wedge (\bm{v}_{2}\wedge \bm{v}_{3})$, it is also deduced that the outer product is associative. Therefore, $r$-blades can be denoted simply as:
 \begin{equation}\label{ibladebis}
 \bm{v}_{1}\wedge \bm{v}_{2} \wedge \dots \wedge \bm{v}_{r}.
 \end{equation}
 Linear combinations of $r$-blades are known as \textit{$r$-vectors}, while linear combinations of $r$-vectors, for $0\leq r\leq n$, are called \textit{multivectors}.

 Applied to the basis elements $\{e_i\}$, the geometric product acts as follows:
 \begin{equation}\label{elementsofbasis}
e_{i}e_{j}  = \left\{\begin{array}{lcl}
 1 & \text{for} & i=j\\
 e_{i}\wedge e_{j} & \text{for} & i\neq j
 \end{array}\right.
 \end{equation} 
 Then, $\{e_{1},\dots,e_{n}\}$ can be expanded to a basis of $\mathcal{G}_{n}$ that contains, for each $0\leq k\leq n$, $C(n,k)$ grade $k$ elements:
 \begin{equation}\label{basisGn}
 \begin{split}
 &\text{Scalar: } 1\\
 &\text{Vectors: } e_{1},\dots,e_{n}\\
 &\text{Bivectors: } \{e_{i}\wedge e_{j}\}_{1\leq i<j\leq n}\\
 &\text{Trivectors: } \{e_{i}\wedge e_{j}\wedge e_{k}\}_{1\leq i<j<k\leq n}\\
 &\vdots\\
 &\text{$n$-blade: } e_{1}\wedge\dots\wedge e_{n}
 \end{split}
 \end{equation}
This allows to extend the geometric product defined in equation \eqref{geometricproduct} to arbitrary multivectors. Readers interested in a detailed explanation of this extension are referred to \cite{DoranLasenby03}. The grade $n$ element $e_{1}\wedge\dots\wedge e_{n}$ is known as the \textit{pseudoscalar} and is usually denoted by $I_{n}$. Pseudoscalars allow us to define one of the main operators of geometric algebra, the \textit{dual} operator. Its action over an $r$-vector $A_{r}$ is:
 \begin{equation}\label{dual}
 A_{r}^{\ast} = I_{n}A_{r},
 \end{equation}
 where $A_{r}^{\ast}$ is an $(n-r)$-vector. In particular, we have that for two multivectors $A$ and $B$, the following identity holds:
 \begin{equation}\label{deMorgan}
 	(A\wedge B)^\ast = A\cdot B^\ast.
 \end{equation}
For $\mathbb{R}^3$, the geometric algebra $\mathcal{G}_3$ has the following basis:
\begin{equation}
\{1,e_{1},e_{2},e_{3},e_{12},e_{13},e_{23},I_{3}\},
\end{equation}
where $e_{ij} = e_{i}\wedge e_{j}$. 

In $\mathcal{G}_3$, bivectors play an important role since they can be used to describe spatial rotations. Indeed, in geometric algebra, rotations are described using \textit{rotors}. If a point $\bm{x}\in\mathbb{R}^{3}$ is rotated by an angle $\theta$ around an axis $\ell$, the rotor $R$ defining such a rotation is:
\begin{equation}\label{RotorDescrip}
R = \cos\biggl(\dfrac{\theta}{2}\biggr)-\sin\biggl(\dfrac{\theta}{2}\biggr)B,
\end{equation}
where $B$ is the unit bivector, i.e., $B^2 = -1$, representing the plane normal to $\ell$. The rotated point $\bm{x}'$ is calculated by sandwiching $\bm{x}$ between $R$ and its reverse $\widetilde{R}$:
\begin{equation}\label{sandwiching}
\bm{x}' = R\bm{x}\widetilde{R}
\end{equation}
where
\begin{equation}
\widetilde{R} = \cos\biggl(\dfrac{\theta}{2}\biggr)+\sin\biggl(\dfrac{\theta}{2}\biggr)B
\end{equation}
and $R\widetilde{R}=1$. Furthermore, equation (\ref{sandwiching}) can be extended to arbitrary multivectors. In addition, to specify what unit multivectors are, a norm is defined in $\mathcal{G}_n$. For an arbitrary multivector $A$, it is:
\begin{equation}
\|A\| = \sqrt{\left<A\widetilde{A}\right>_0},
\end{equation}
where $\left<\cdot\right>_0$ is the grade-0 projection operator, i.e.,  it extracts the scalar elements of the argument multivector.

It will be seen in the following subsection that translations can also be described by {\em rotors} in the {\em conformal geometric algebra}, a five-dimensional representation of three-dimensional Euclidean space. In addition, general geometric entities (such as points, lines, planes, circles, etc.) and the relations between them are also easily described in conformal geometric algebra.

\subsubsection{Conformal geometric algebra}\label{CGA}
The conformal model extends the real vector space $\mathbb{R}^{3}$ by adding two extra basis vectors, $e$ and $\overline{e}$, with the property:
\begin{equation}
e^{2}=1,\quad\overline{e}^{2}=-1.
\end{equation} 
The enlarged vector space is usually denoted by $\mathbb{R}^{4,1}$. In addition, these two extra vectors allow us to define two null vectors, i.e., vectors whose square vanishes:
\begin{equation}
n_0=\dfrac{1}{2}\left(\overline{e}-e\right),\quad n_\infty=\overline{e}+e,
\end{equation}
where $n_\infty$ is associated with the point at infinity and $n_0$ with the origin. However, in order to avoid confusion with the number of DoF of the robot, denoted by $n$, an alternative yet accepted notation will be used throughout the rest of the paper:
\begin{equation}\label{nullvectors}
	e_0=\dfrac{1}{2}\left(\overline{e}+e\right),\quad e_\infty=\overline{e}-e,
\end{equation}
where, again, $e_\infty$ is associated with the point at infinity and $e_0$ with the origin,

Thus, the conformal geometric algebra of $\mathbb{R}^{3}$  is denoted by $\mathcal{G}_{4,1}$ and can be seen as a geometric algebra of higher dimension (specifically, the geometric algebra of $\mathbb{R}^{4,1}$). Now, the two null vectors $e_\infty$ and $e_0$ allow an intuitive description of translations, that are formulated through \textit{translators} as follows:
\begin{equation}\label{translatorCGA}
T_{\bm{v}} = 1 - \dfrac{\bm{v}e_\infty}{2},
\end{equation}
where the translation is performed in the direction of $\bm{v}\in\mathbb{R}^{3}$. Since $(\bm{v}e_\infty)^{2}= 0$, translators can be seen as rotors in which the bivector squares to zero. Thus, the translation of an element $A\in\mathcal{G}_{4,1}$ is done as with spatial rotations:
\begin{equation}\label{sandwiching1}
A' = T_{\bm{v}}A\widetilde{T}_{\bm{v}},
\end{equation}
where:
\begin{equation}
	\widetilde{T}_{\bm{v}} = 1 + \dfrac{\bm{v}e_\infty}{2}.
\end{equation}
One of the most important advantages of conformal geometric algebra is that it provides a homogeneous model for the $n$-dimensional Euclidean space. In particular, every point $\bm{x}\in\mathbb{R}^{3}$ is associated with a null vector of $\mathcal{G}_{4,1}$ (including the origin and the point at the infinity). This is done via the Hestenes' embedding:
\begin{equation}\label{Hestenes}
X = H(\bm{x}) = \dfrac{1}{2}\bm{x}^2e_\infty+e_0 +\bm{x},
\end{equation}
where $X$ is said to be the \textit{null vector representation} of $\bm{x}$. Now, we can easily deduce the following relation \cite{Dorst-Fontijne-Mann-2007}:
\begin{equation}\label{distance}
X_{1}\cdot X_{2} = -\dfrac{1}{2}d(\bm{x}_{1},\bm{x}_{2})^2,
\end{equation}
where $X_{1}, X_{2}$ are the null vector representations of $\bm{x}_{1},\bm{x}_{2}\in\mathbb{R}^3$ and $d(\cdot,\cdot)$ denotes the Euclidean distance. As a consequence, the distance between two null vectors of $\mathcal{G}_{4,1}$ can be defined via the Euclidean distance:
\begin{equation}\label{distance_cga}
	d(X_1,X_2) = \sqrt{-2(X_1\cdot X_2)}.
\end{equation}
The relation \eqref{distance} gives us a way to define geometric objects as elements of $\mathcal{G}_{4,1}$. Let $O$ be a geometric object, then $k$ is said to be the \textit{inner representation} of $O$ if  for every point $\bm{x}\in O$, $H(\bm{x}) =X\cdot k= 0$. Now, using the relation \eqref{deMorgan}, we have that $0 = (X\cdot k)^\ast = X\wedge k^\ast$ so $k^\ast$ is also a representation of $O$. In this case, we say that $k^\ast$ is the \textit{outer representation} of $O$ and, in order to avoid confusion, we will denote it by $K$.

Now, for two different geometric objects $O_1$ and $O_2$ with outer (inner) representations $K_1$ and $K_2$ ($k_1$ and $k_2$), we define its \textit{meet} or \textit{intersection} $k_1\vee k_2$ as the multivector 
\begin{equation}\label{intersection}
	k_1\vee k_2 = (k_1\wedge k_2)^\ast = k_1\cdot K_2.
\end{equation}
Analogously, if the outer representations of $O_1$ and $O_2$ have the same grade, the angle defined by them is computed as follows:
\begin{equation}\label{angle}
	\angle(O_1,O_2) = \cos^{-1}\left(\dfrac{K_1\cdot K_2}{|K_1||K_2|}\right).
\end{equation}
To describe the geometric objects we are going to use throughout this work, let $\bm{p}_1,\bm{p}_2,\bm{p}_3,\bm{p}_4\in\mathbb{R}^3$ be four different points with null vector representation $P_1,P_2,P_3,P_4\in\mathcal{G}_{4,1}$. Then:
\begin{itemize}
	\item $B = P_i\wedge P_j$ (with $1\leq i\neq j \leq 4$) is a bivector and the outer representation of the pair of points $\bm{p}_i$ and $\bm{p}_j$.
	\item $L = P_i\wedge P_j\wedge e_\infty$ (with $1\leq i\neq j \leq 4$) is a trivector and the outer representation of the line with direction $\bm{p}_i$ to $\bm{p}_j$. Its inner representation is the bivector $\ell= \bm{v}e_{123} - (\bm{p}_i\wedge \bm{v})e_{123}e_\infty$, where $\bm{v} = \bm{p}_j-\bm{p}_i$ is its direction vector.
	\item $C = P_i\wedge P_j\wedge P_k$ (with $1\leq i\neq j\neq k\leq 4$) is a trivector and the outer representation of a circle passing through the points $\bm{p}_i, \bm{p}_j$ and $\bm{p}_k$. Its inner representation is the bivector $c = \pi\wedge s$, where $\pi$ and $s$ are the inner representations of the plane and sphere whose intersection defines the circle.
	\item $\Pi = P_i\wedge P_j\wedge P_k\wedge e_\infty$ (with $1\leq i\neq j\neq k\leq 4$) is a 4-vector and the outer representation of a plane passing through the points $\bm{p}_i, \bm{p}_j$ and $\bm{p}_k$. Its inner representation is the vector $\pi = \bm{n} + \delta e_\infty$, where $\bm{n}$ denotes the vector normal to the plane and $\delta$, its orthogonal distance to the origin.
	\item $S = P_1\wedge P_2\wedge P_3\wedge P_4$ is a 4-vector and the outer representation of a sphere passing through the points $\bm{p}_1, \bm{p}_2,\bm{p}_3$ and $\bm{p}_4$. Its inner representation is the vector $s = Z - \frac{1}{2}r^2e_\infty$, where $Z$ is the null vector representation of the center of the sphere and $r$, its radius.
\end{itemize}

\section{A conformal geometric algebra formulation of forward kinematics}\label{FK}
The conformal model of the three-dimensional Euclidean space $\mathcal{G}_{4,1}$ provides an elegant and compact way of describing the forward kinematics. Contrary to the classical approach, where we use the homogeneous transformation matrices constructed with the D-H parameters, this approach is entirely based on the use of rotors.

As stated in section \ref{Intro}, each joint frame is described with respect to the preceding one, i.e., the $i$-th joint frame is constructed from the $(i-1)$-th joint frame by the successive application of rigid motions (translations and rotations). Since both are described in $\mathcal{G}_{4,1}$ by rotors, for each joint $i$, the following rotors are defined:
\begin{equation}\label{RotorsDH}
\begin{split}
T_{d_{i}} &= 1 - d_{i}\dfrac{\bm{z}_{i}e_\infty}{2},\\
R_{\theta_{i}} &= \cos\biggl(\frac{\theta_{i}}{2}\biggr)-\sin\biggl(\frac{\theta_{i}}{2}\biggr)\bm{x}_{i}\wedge\bm{y}_{i},\\
T_{a_{i}} &= 1 - a_{i}\dfrac{\bm{x}_{i-1}e_\infty}{2},\\
R_{\alpha_{i}} &= \cos\biggl(\frac{\alpha_{i}}{2}\biggr)-\sin\biggl(\frac{\alpha_{i}}{2}\biggr)\bm{y}_{i-1}\wedge\bm{z}_{i-1}.
\end{split}
\end{equation}
where $\{\bm{x}_{i},\bm{y}_{i},\bm{z}_{i}\}$ (resp. $\{\bm{x}_{i-1},\bm{y}_{i-1},\bm{z}_{i-1}\}$) denotes the three basis elements of the $i$-th (resp. $(i-1)$-th) joint frame. Then, two main rotors can be constructed from the previous ones:
\begin{equation}\label{RotorsScrewDH}
\begin{split}
M_{\theta_{i}} &= T_{d_{i}}R_{\theta_{i}},\\
M_{\alpha_{i}} &= T_{a_{i}}R_{\alpha_{i}}.
\end{split}
\end{equation}
Notice that rotor $M_{\theta_{i}}$ contains both joint variables, while $M_{\alpha_{i}}$ is a constant rotor. $M_{\theta_i}$ represents a screw motion around the $i$-th joint axis, $\bm{z}_i$, while $M_{\alpha_{i}}$ represents a screw motion around the $\bm{x}_{i-1}$ axis of the $(i-1)$-th joint frame.

Now, we can formulate the forward kinematics of a serial robot of $n$ DoF as follows:
\begin{equation}\label{ForwardRotorEquation}
X' = M_{\theta_{1}}M_{\alpha_{1}}\cdots M_{\theta_{n}}M_{\alpha_{n}}X\widetilde{M}_{\alpha_{n}} \widetilde{M}_{\theta_{n}}\cdots\widetilde{M}_{\alpha_{1}}\widetilde{M}_{\theta_{1}},
\end{equation}
where $X'$ ($X$) is the null vector representation of either the end-effector (reference frame) position vector or the vectors defining the end-effector (reference frame) orientation. We can compactly rewrite identity \eqref{ForwardRotorEquation} as follows:
\begin{equation}
	X'=M_{1}(q_1)\cdots M_{n}(q_n)X\widetilde{M}_{n}(q_n)\cdots\widetilde{M}_{1}(q_1) = M(\bm{q})X\widetilde{M}(\bm{q}),
\end{equation}
where $M(\bm{q}) = M_{1}(q_1)\cdots M_{n}(q_n)$ and $M_i(q_i) = M_{\theta_{i}}M_{\alpha_{i}}$. Basically, what we are doing is to transform the fixed reference frame (usually placed at the base of the robot) into the end-effector frame according to configuration $\bm{q}$.

\section{Pose representation in conformal geometric algebra}\label{RPCGA}
If the desired end-effector pose is represented in matrix form as: 
\begin{equation}
	T = \begin{pmatrix}
		R & \bm{p}\\
		0 & 1
	\end{pmatrix},
\end{equation}
then it is possible to construct a rotor $M\in\mathcal{G}_{4,1}$ that also represents such a pose. Indeed, given the matrix representation of the reference frame:
\begin{equation}
T_{0} = \begin{pmatrix}
I_{3} & 0\\
0 & 1
\end{pmatrix},
\end{equation}
we can compute the rotor that transforms $T_0$ into $T$. First, the position vector $\bm{p}$ defines the rotor:
\begin{equation}\label{translator}
T_{\bm{p}} = 1 - \|\bm{p}\|\dfrac{\bm{p}e_\infty}{2},
\end{equation}
that applied to $e_0$ give us $P = e_0 + \bm{p} + \frac{1}{2}\bm{p}^2e_\infty$, i.e., the null vector representation of $\bm{p}$.
Now, to compute the rotor that relates the orientations determined by the rotations matrices $R$ and $I_{3}$, we notice that both rotation matrices can be seen as two sets of orthogonal vectors in $\mathbb{R}^3$, namely $\{e_{1},e_{2},e_{3}\}$ and $\{f_{1},f_{2},f_{3}\}$. Then, the problem is reduced to find a rotor transforming one set of orthogonal vectors into another. In \cite[pag. 103]{DoranLasenby03}, a simple way of computing such rotor is presented. It involves the use of the \textit{reciprocal frame}. Associated with any arbitrary set of orthogonal vectors $\{e_{1},\dots,e_{n}\}$, there exists another set of orthogonal vectors, denoted by $\{e^{1},\dots,e^{n}\}$ and defined by the property:
\begin{equation}\label{reciprocalframe}
e^{i}\cdot e_{j} =\delta_{ij}\,\text{ for all } i,j = 1,\dots,n,
\end{equation}
where $\delta_{ij}$ denotes the Kronecker $\delta$. Such a set is said to be the reciprocal frame of $\{e_{1},\dots,e_{n}\}$. 

Now, following \cite[pag. 103]{DoranLasenby03}, the following formula is established:  
\begin{equation}\label{Rotor}
R_{I_3,R} = \dfrac{1 + f_{1}e^{1}+f_{2}e^{2}+f_{3}e^{3}}{|1 + f_{1}e^{1}+f_{2}e^{2}+f_{3}e^{3}|},
\end{equation}
where, again, $\{e_1, e_2, e_3\}$ is the orientation of the reference frame, while $\{f_1, f_2, f_3\}$ is the desired orientation of the end-effector. Finally, the rotor $M$ that transforms $T_{0}$ into $T$ is the product of rotors (\ref{translator}) and (\ref{Rotor}):
\begin{equation}\label{RotorInput}
M = T_{\bm{p}}R_{I_3,R}.
\end{equation}
As stated in section \ref{Intro}, for any serial robot with a spherical wrist, the target position $\bm{p}$ can be moved to the wrist center point $\bm{p}_{w}$ by a fixed transformation. Then, the new target position is $\bm{p}_{w}$ and thus, it can be assured that the first three joints contribute to the position and orientation, while the last three only contribute to the orientation. This allows us to decouple the inverse kinematics into the position and the orientation subproblems.

\section{Solutions for non-redundant robots with spherical wrist}\label{NRRSW}
In this section, the position and orientation problems for a 6 DoF serial robot with a spherical wrist are solved. Again, for the position problem different geometric strategies are defined depending on the nature and disposition of the joints along the kinematic chain. For the orientation problem, the rotor $R_{I_3,R}$ is split into three different rotors from which the joint variables can be obtained.

\begin{figure*}[t!]
	\centering
	\begin{subfigure}[b]{0.45\textwidth}
		\includegraphics[width=0.85\textwidth]{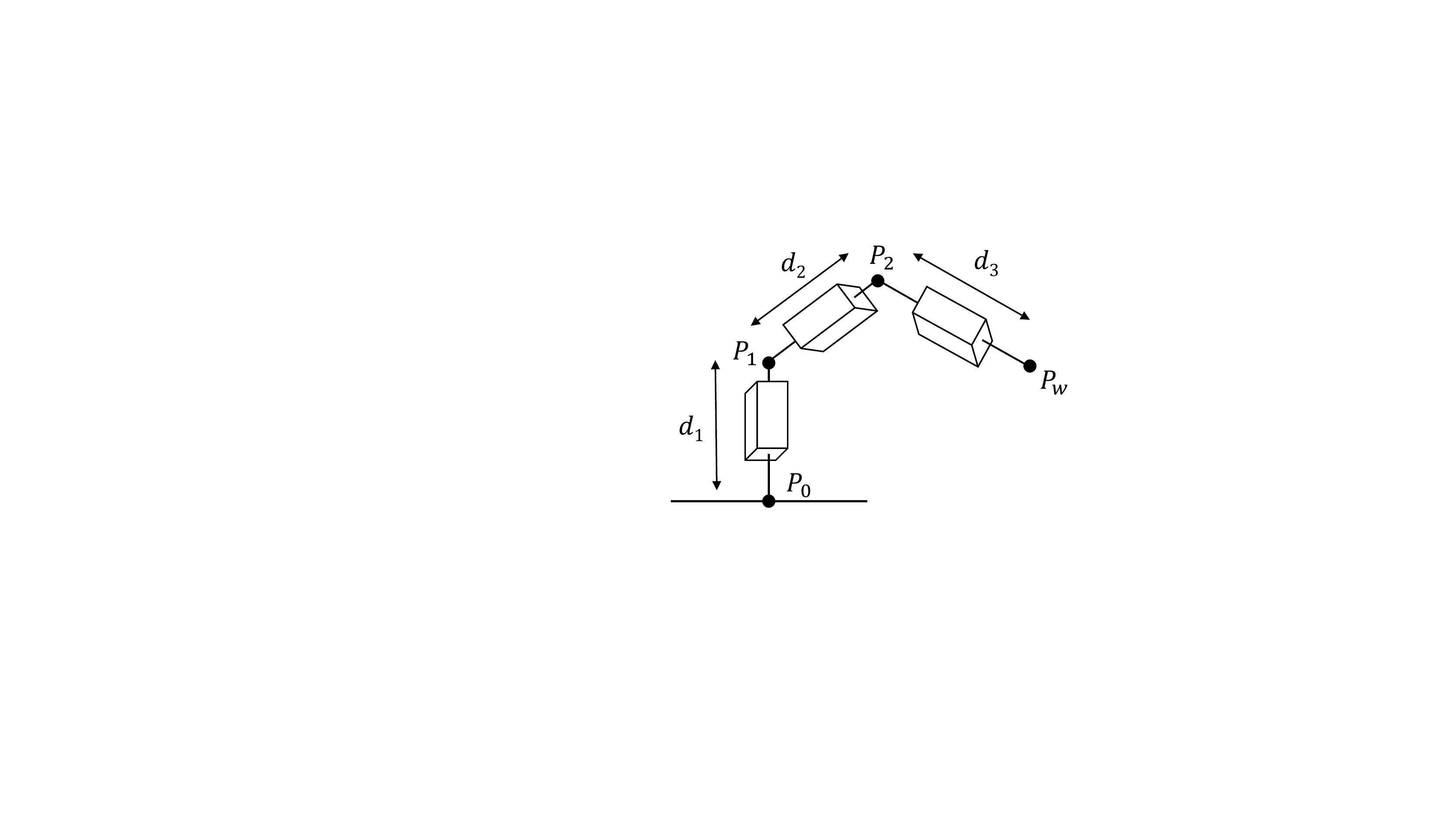}	
		\caption{P+P+P}
		\label{PPP}
	\end{subfigure}
	\begin{subfigure}[b]{0.45\textwidth}
		\includegraphics[width=0.85\textwidth]{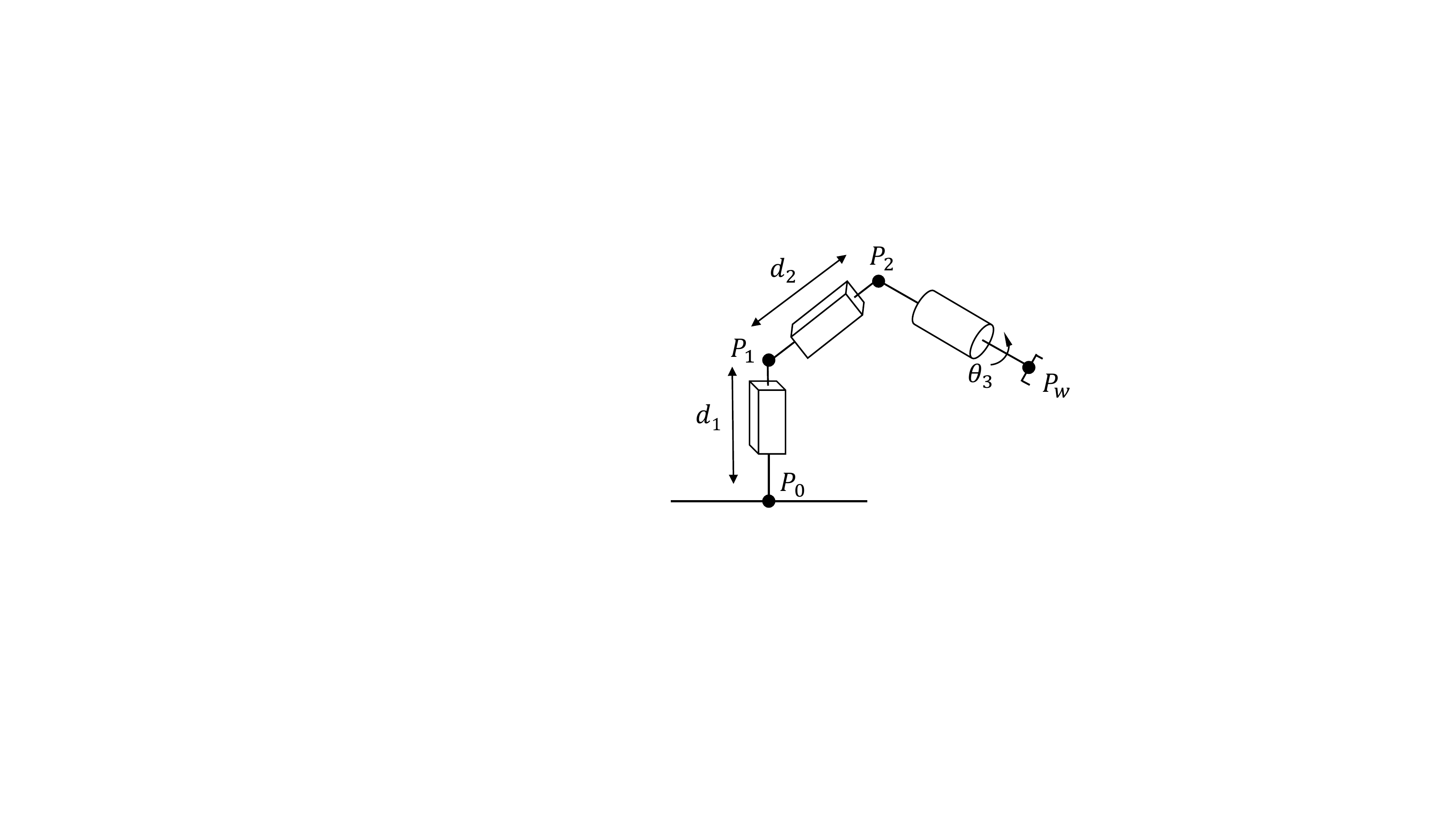}	
		\caption{P+P+R}
		\label{PPR}
	\end{subfigure}
	\begin{subfigure}[b]{0.45\textwidth}
		\includegraphics[width=0.85\textwidth]{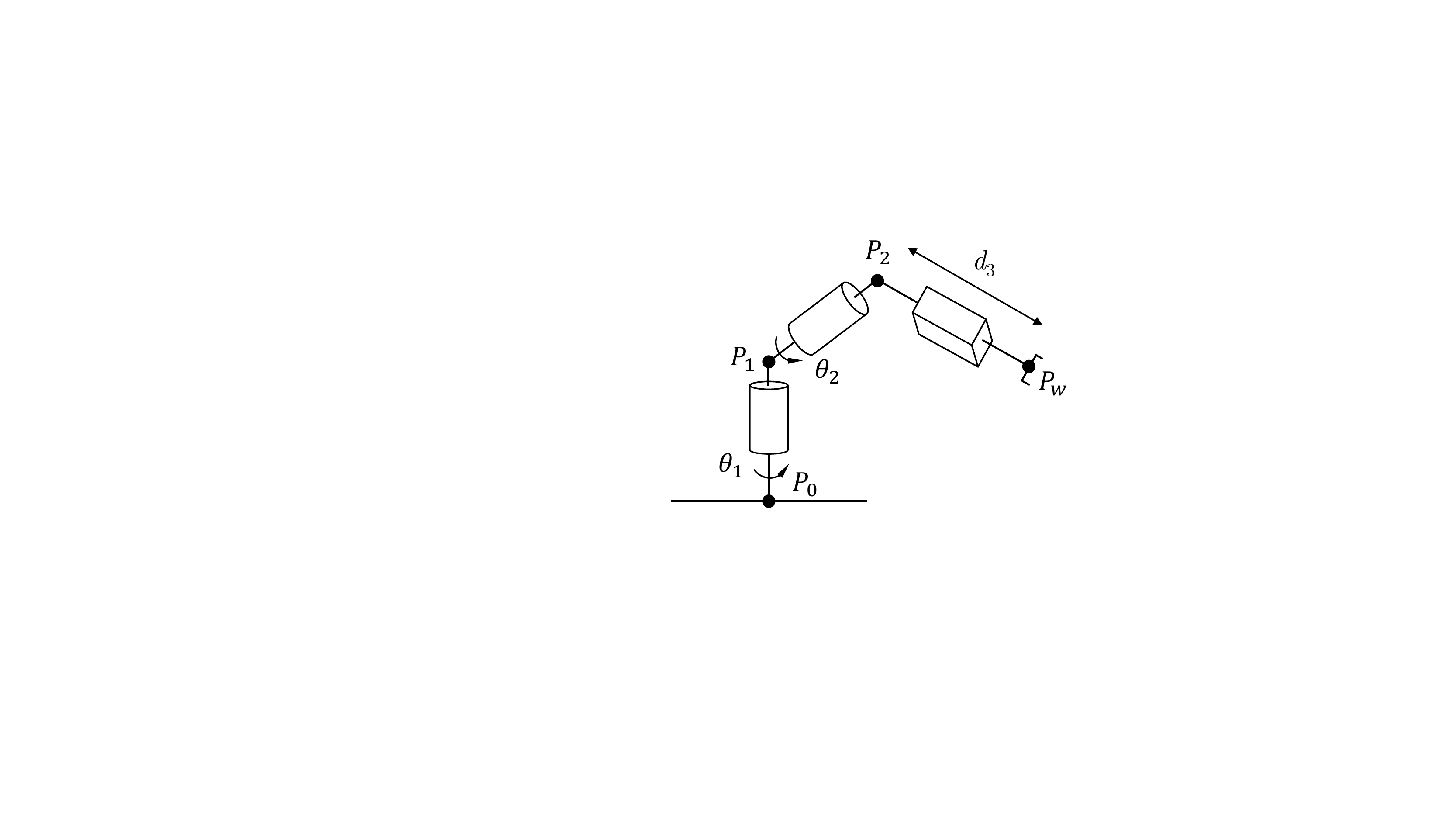}	
		\caption{R+R+P}
		\label{RRP}
	\end{subfigure}
	\begin{subfigure}[b]{0.45\textwidth}
		\includegraphics[width=0.85\textwidth]{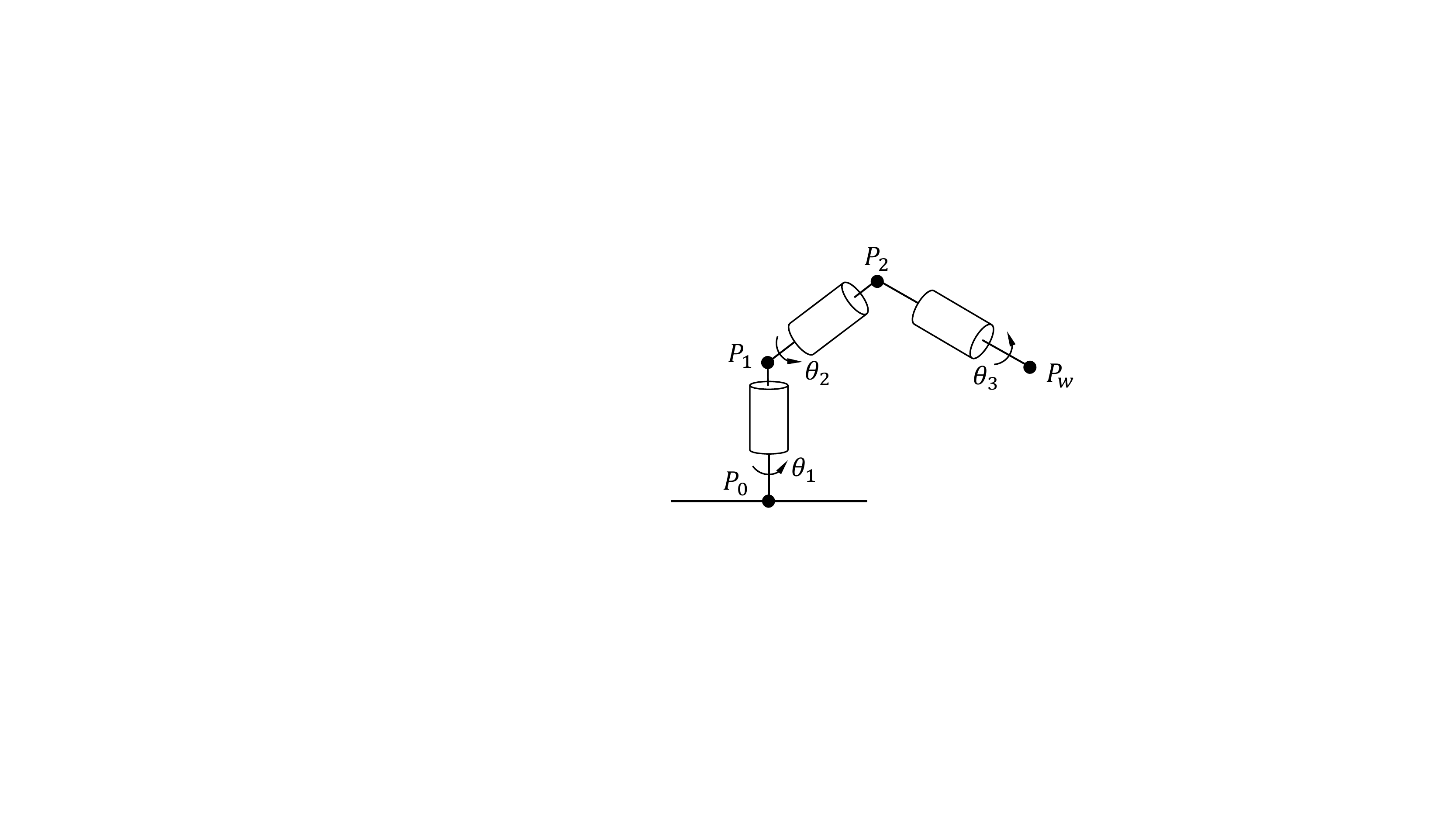}
		\caption{R+R+R}
		\label{RRR}
	\end{subfigure}
	\caption{Different combinations of the joints forming the position part of a serial robot with a spherical wrist. Here, R denotes a revolute joint and P denotes a prismatic joint.}
	\label{PositionPart}
\end{figure*}

\subsection{Position problem}\label{SPP}
Four different combinations of prismatic and revolute joints that constitute the position part of the robot are analysed (figure \ref{PositionPart}). For each of them, a geometric strategy is developed. It consists of computing the null vector representation of some auxiliary points placed at each joint to recover the corresponding joint variables as the angle or displacement between two geometric objects defined with such points. For each case, the point at the origin, $e_0$, is denoted by $P_{0}$ and is placed at the base of the robot. Similarly, the target position $\bm{p}_{w}$ is expressed as a null vector, $P_{w}$, obtained through the Hestenes' embedding (\ref{Hestenes}). Finally, depending on the information available, for a given geometric object it could be convenient to use its inner representation $k$ or its outer representation $K$. Clearly, a geometric object described with one of these two representations can be expressed in the other representation by using the dual operator (as explained in section \ref{CGA}):
\begin{equation}
K = I_5k.
\end{equation}

\subsubsection{Three prismatic joints (P+P+P)}\label{InPoPPP}
A scheme of the position part of a serial robot with three prismatic joints is depicted in figure \ref{PPP}. First of all, the case without offsets between consecutive joints is considered, i.e., the axes of each pair of consecutive joints intersect. Given the first translation axis $\bm{z}_{1}$, two of the rotors defined in (\ref{RotorsDH}) can be used to obtain the joint axes $\bm{z}_{2}$ and $\bm{z}_{3}$ as follows:
\begin{equation}\label{z2z3PPP}
\begin{split}
\bm{z}_{2} &= R_{\theta_2}R_{\alpha_{2}}\bm{z}_{1}\widetilde{R}_{\alpha_{2}}\widetilde{R}_{\theta_2},\\
\bm{z}_{3} &= R_{\theta_3}R_{\alpha_{3}}\bm{z}_{2}\widetilde{R}_{\alpha_{3}}\widetilde{R}_{\theta_3}.
\end{split}
\end{equation}

The joint axis $\bm{z}_{3}$ together with the target position $\bm{p}_{w}$ defines a line whose inner representation is:
\begin{equation}\label{l3PPP}
\ell_{3} = \bm{z}_{3}e_{123} - (\bm{p}_{w}\wedge\bm{z}_{3})e_{123}e_\infty.
\end{equation}
Now, a plane containing the joint axes $\bm{z}_{1}$ and $\bm{z}_{2}$ and passing through the point $P_{0}$ is also defined so its inner representation is:
\begin{equation}\label{plano1PPP}
\pi_{1}= \bm{z}_{1}\times\bm{z}_{2},
\end{equation}
where $\times$ denotes the cross product between three-dimensional vectors.

\begin{figure*}[t!]
	\centering
	\begin{subfigure}[b]{0.45\textwidth}
			\includegraphics[width=7cm]{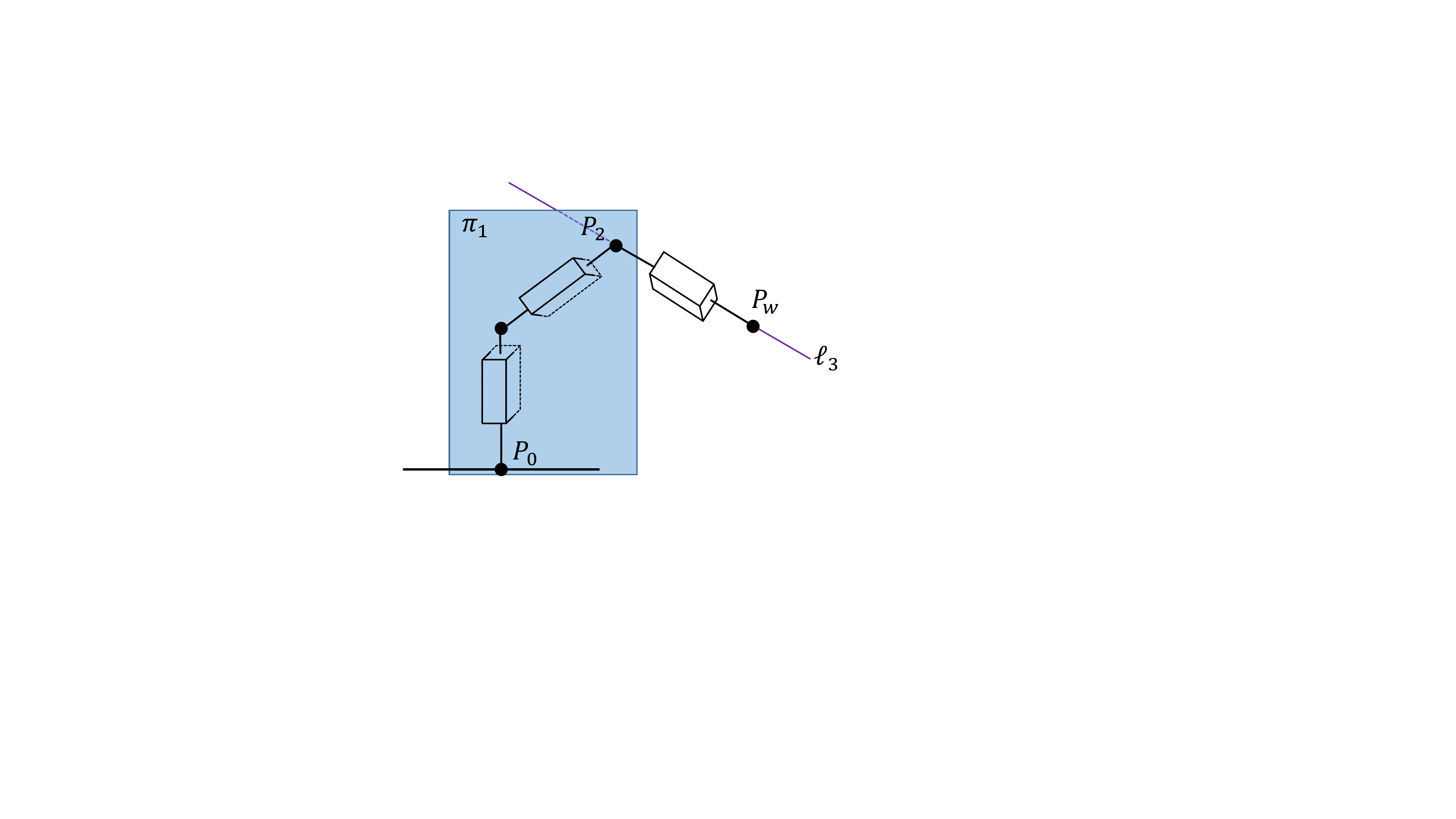}
			\hspace{0.2cm}
			\caption{Computation of null vector $P_2$ as the intersection point of $\ell_3$ and $\pi_1$.}
			\label{PPP_2_1}
	\end{subfigure}
	\hspace{0.5cm}
	\begin{subfigure}[b]{0.45\textwidth}
			\includegraphics[width=6cm]{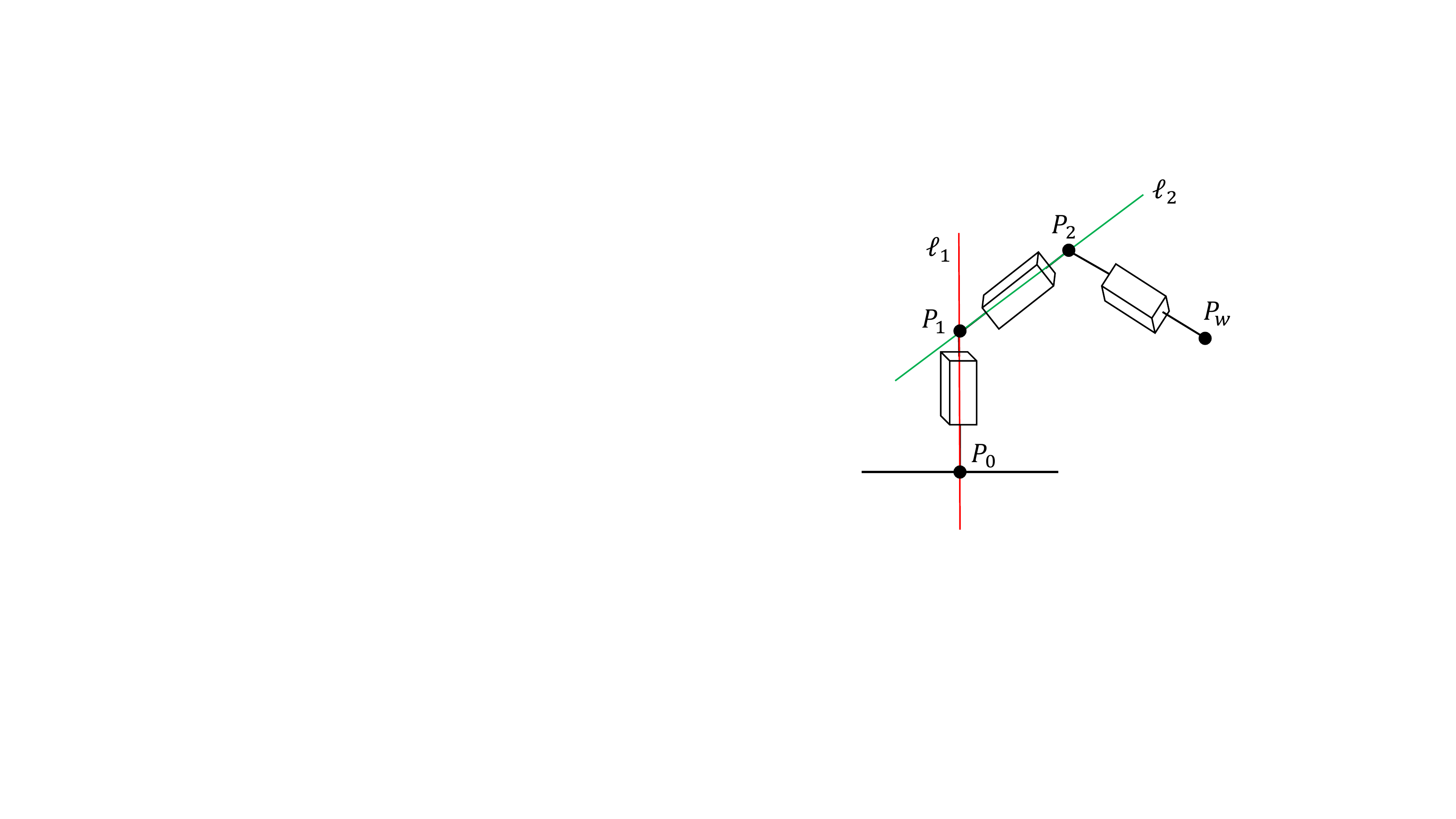}
			\caption{Computation of null vector $P_1$ as the intersection point of $\ell_1$ and $\ell_2$.}
			\label{PPP_2_2}
	\end{subfigure}
	\caption{Geometric computation of the null vectors $P_2$ and $P_1$.}
\end{figure*}

Since the end-effector position is not restricted to a fixed plane, there are not parallel prismatic joint axes. Moreover, $\bm{z}_1\wedge\bm{z}_2\wedge\bm{z}_3\neq 0$ and, thus, the intersection of the line and plane represented by $\ell_{3}$ and $\pi_{1}$ is non-empty (as graphically shown in figure \ref{PPP_2_1}):
\begin{equation}\label{P2PPP}
B = \ell_{3}\vee \pi_{1},
\end{equation}
where $B\neq 0$ is a bivector (as deduced from identity (\ref{intersection})). Hence, it represents a pair of points in conformal geometric algebra. However, since the intersection between a line and a plane is a single point, the bivector $B$ is of the form:
\begin{equation}
	B = P_{2}\wedge e_\infty
\end{equation}
for a null point $P_{2}$, that can be extracted from $B$ as explained in \cite[pags. 24-26]{Aristidou10}. With this null point, the following lines are defined through their inner representations:
\begin{equation}\label{l1l2}
\begin{split}
\ell_{1} &= \bm{z}_{1}e_{123}-(e_0\wedge\bm{z}_{1})e_{123}e_\infty,\\
\ell_{2} &= \bm{z}_{2}e_{123}-(\bm{p}_{2}\wedge\bm{z}_{2})e_{123}e_\infty,
\end{split}
\end{equation}
where $\bm{p}_{2}$ is the vector whose null vector representation is $P_{2}$ and that is recovered with the projection $P:\mathbb{R}^{4,1}\to\mathbb{R}^3$, that is, the inverse of the Hestenes' embedding \eqref{Hestenes}. In addition, if $L_1,L_2$ and $\Pi_1$ are the outer representations of the two lines  and plane defined before, $L_{1}\wedge\Pi_{1}=0$ and $L_{2}\wedge\Pi_{1}=0$ and, since the joint axes $\bm{z}_{1}$ and $\bm{z}_{2}$ are not parallel, they have non-empty intersection. This can be seen depicted in figure \ref{PPP_2_2}. However, this intersection is a special case. Indeed
\begin{equation}\label{P1PPP}
\ell_{1}\vee \ell_{2} = \left(\ell_{1}\wedge \ell_{2}\right)^{\ast}
\end{equation}
is a grade one element, i.e., a vector. Non-null vectors do not represent any geometric object in conformal geometric algebra. Therefore, we need an alternative method for obtaining the intersection point. In this paper, we follow the same technique as in \cite[pag. 31]{Aristidou10} to compute $P_1 = \ell_{1}\vee \ell_{2}$.

Finally, with the null vectors $P_{0},P_{1},P_{2}$ and $P_{w}$, the joint variables are recovered using the distance (\ref{distance_cga}):
\begin{equation}\label{d1d2d3}
d_{1} = d(P_{0},P_{1}), d_{2} = d(P_{1},P_{2}) \text{ and }d_{3} = d(P_{2},P_{w}).
\end{equation}
Now, the case where there is an offset between two consecutive joints is studied. Offsets are always modeled by the four D-H parameters $a$, $\alpha$, $\theta$ and $d$. If joint $i$ is revolute (prismatic), its joint variable can have an offset as well (aligned with the joint axis by definition). In that case, we write $\theta_i = \hat{\theta}_i + \overline{\theta}_i$ ($d_i = \hat{d}_i + \overline{d}_i$), where $\hat{\theta}_i$ ($\hat{d}_i$) denotes the real joint variable and $\overline{\theta}_i$ ($\overline{d}_i$), the length of the offset.

\begin{figure*}[t!]
	\centering
	\begin{subfigure}[b]{0.45\textwidth}
		\includegraphics[width=5cm]{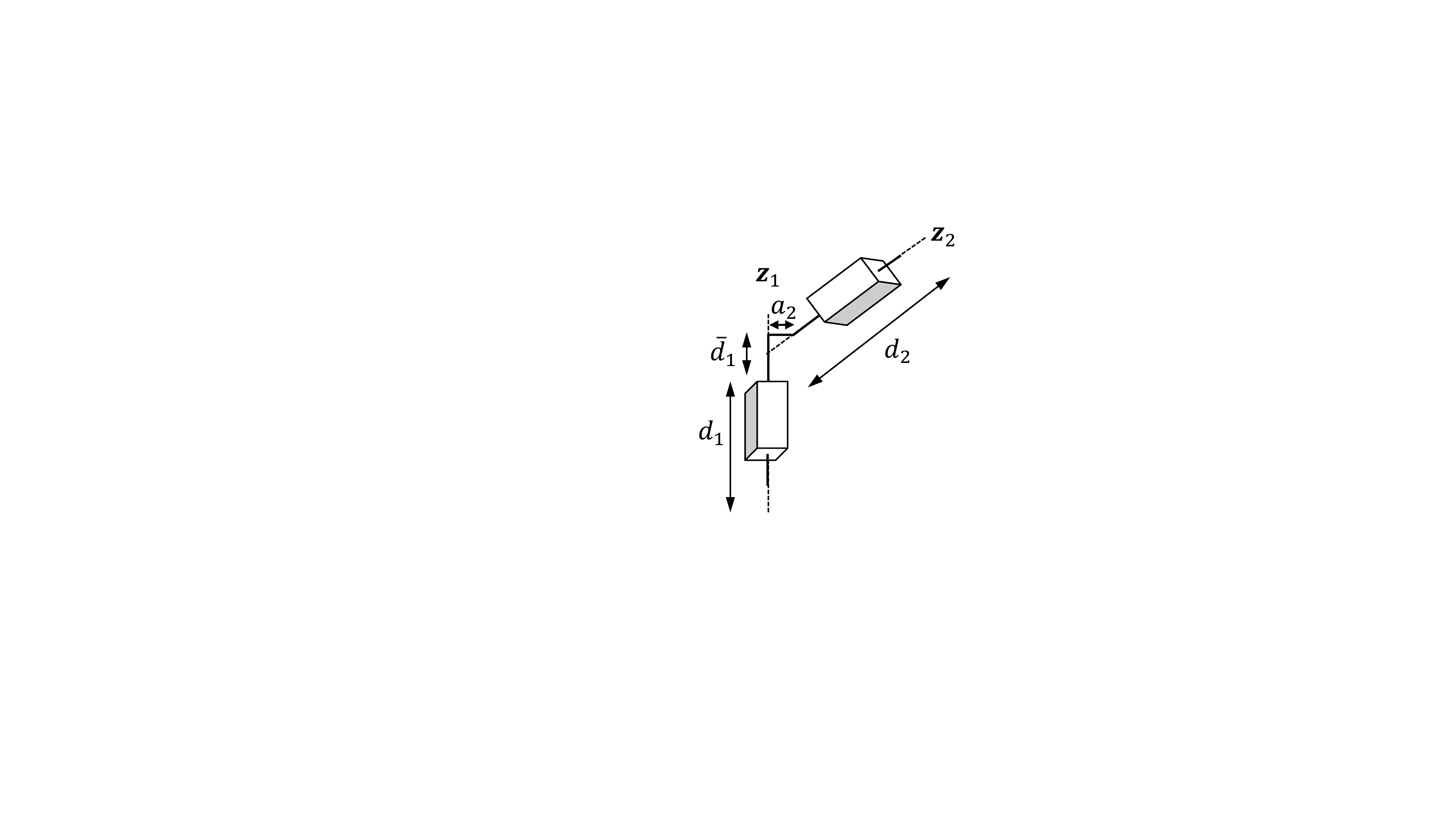}
		\caption{Example of an offset between the first two joint axes.}
		\label{PPPoffset}
	\end{subfigure}
	\hspace{0.5cm}
	\begin{subfigure}[b]{0.45\textwidth}
		\includegraphics[width=4cm]{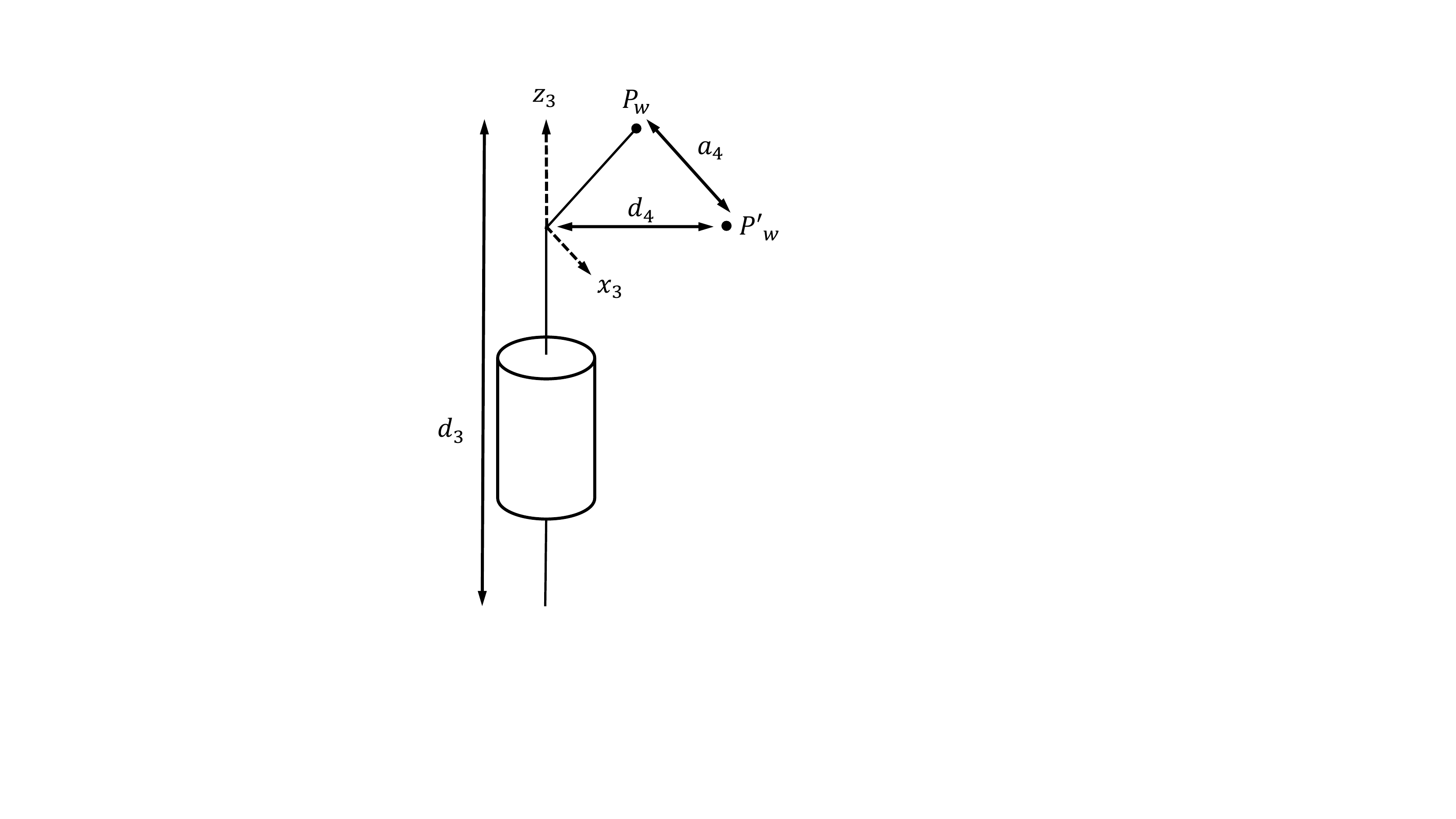}
		\caption{Translation of $P_{w}$ to compensate the offset placed at the end of the third link.}
		\label{offsetPw}
	\end{subfigure}
	\caption{Offsets in the position part of serial robots with spherical wrist.}
\end{figure*}

Let us suppose, without loss of generality, that there exists an offset of fixed length between the first two joints (the other cases are analogous). Clearly, such offset is either aligned with the common perpendicular between the first and second joint axes and has length $a_2$ or can be decomposed into two components (as depicted in figure \ref{PPPoffset}): one aligned with such a  common perpendicular (with length $a_2$) and the other one aligned with the first joint axis (with length $\overline{d}_1$). Now, since the plane represented by $\pi_{1}$ is defined with the joint axes $\bm{z}_{1}$ and $\bm{z}_{2}$ and, in this case, offsets do not change the direction of the joint axes, we can translate it in the direction of the common perpendicular by an amount equal to $a_2$:
\begin{equation}
\pi'_1= T_{a_2}\,\pi_{1}\,\widetilde{T}_{a_2},
\end{equation}  
where $T_{a_2}$ is defined as in \eqref{RotorsDH}. Now, $P_2\in\pi'_1$ and, therefore, $P_{2}$ can be obtained from equation (\ref{P2PPP}) as in the case without offsets. With $P_{2}$, the inner representation $\ell_{2}$ of a line is determined as in (\ref{l1l2}). Then, this line is transformed as follows:
\begin{equation}\label{ell2bis}
\ell'_{2} = T_{-\overline{d}_1}T_{-a_2}\,\ell_{2}\,\widetilde{T}_{-a_2}\widetilde{T}_{-\overline{d}_1},
\end{equation}
where, again, $T_{-a_2}$ and $T_{-\overline{d}_1}$ are defined as in \eqref{RotorsDH}. Now, since the lines represented by $\ell_{1}$ and $\ell'_{2}$ belong to the same plane, they have non-empty intersection and $P_{1}=\ell_{1}\vee\ell'_{2}$. This completes the solution for this case. 

%Offsets between consecutive joints are always modeled by the four D-H parameters: $a,\alpha,\theta$ and $d$. If joint $i$ is revolute (prismatic), then $\theta_{i}$ ($d_{i}$) is a joint variable and, therefore, it is decomposed into $\theta_{i} = \widehat{\theta}_{i}+\overline{\theta}_{i}$ ($d_{i} = \widehat{d}_{i} + \overline{d}_{i}$) where $\widehat{\theta}_{i}$ ($\widehat{d}_{i}$) is a constant value, usually zero, that describes the component of the offset aligned with the joint axis, while $\overline{\theta}_{i}$ ($\overline{d}_{i}$) corresponds to the real joint variable and can be renamed as the original one (i.e., from now on, $\overline{\theta}_{i}$ and $\overline{d}_{i}$ are denoted by $\theta_{i}$ and $d_{i}$). Hence, an offset between the joints $i-1$ and $i$ can be described by $(a_{i},\alpha_{i},\widehat{\theta}_{i},d_{i})$ if joint $i$ is revolute or $(a_{i},\alpha_{i},\theta_{i},\widehat{d}_{i})$ if it is prismatic. In either of these two cases, the offset can be compensated simply by applying a sequence of rotors to the proper geometric entity.

\subsubsection{Two prismatic joints and one revolute (P+P+R)}\label{InPoPPR}
The position part of a serial robot with two prismatic joints and one revolute is depicted in figure \ref{PPR}. There are different combinations of two prismatic joints with one revolute but most of them are treated in a similar way. Only the most relevant cases are fully developed here. Again, the case without offsets is considered first.

Let us suppose that the first two joints are prismatic and the third one, revolute. As in the preceding case, given $\bm{z}_{1}$, the joint axes $\bm{z}_{2}$ and $\bm{z}_{3}$ can be calculated as in (\ref{z2z3PPP}). Now, the inner representation of the plane that contains $P_{0}$ and the joint axes $\bm{z}_{1}$ and $\bm{z}_{2}$ is defined as in (\ref{plano1PPP}) and denoted by $\pi_{1}$. Since the robot has 6 DoF, its end-effector position is not restricted to a fixed plane and, thus, there exists an offset at the end of the third link, i.e., between the third link and the set of joints that forms the spherical wrist. In addition, $\bm{z}_{3}$ cannot be orthogonal to the plane represented by $\pi_{1}$ (if it was, $\bm{p}_{w}$ would always belong to a fixed plane parallel to the plane represented by $\pi_{1}$). 

\begin{figure*}[t!]
	\centering
	\includegraphics[width=10cm,height=6cm]{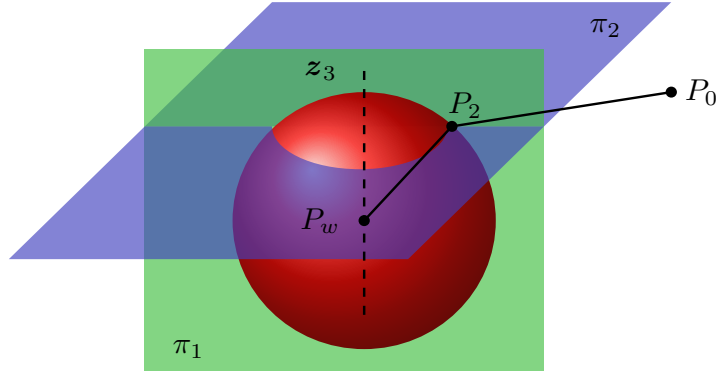}
	\caption{$P_{2}$ as the intersection of two planes and one sphere.}
	\label{Spheres+plane}
\end{figure*}

Since the last three joints of the robot do not contribute to the position of the end-effector, they are not considered in the position problem. Therefore, this offset can be seen as a displacement defined in the $\bm{x}$--$\bm{z}$ plane of the third joint frame, i.e., a displacement of length $a_{4}$ in the $\bm{x}_{3}$ direction. Then, it is possible to translate $P_{w}$ to compensate the offset (as depicted in figure \ref{offsetPw}):
\begin{equation}\label{P'w}
P'_{w} = T_{-a_4}P_{w}\widetilde{T}_{-a_4},
\end{equation}
where $T_{-a_4}$ is defined as in \eqref{RotorsDH}. Now, the null point $P_{2}$ belongs to the intersection of two planes with one sphere. One of these planes is represented by $\pi_{1}$, while the other is defined as follows:
\begin{equation}\label{pi2PPR}
\pi_{2} = \bm{z}_{3} + d(P_{0},P'_{w})e_\infty.
\end{equation}
Additionally, the inner representation of the sphere is defined as:
\begin{equation}\label{S1PPR}
s_{1} = P'_{w} - \dfrac{1}{2}(d_{3}^{2}+d_{4}^{2})e_\infty,
\end{equation}
where $d_{3}$ denotes the length of the third link and $d_4$, the distance between the origin of the third frame and $\bm{p}_w$ measured in the direction of the $\bm{z}_4$ axis. Since the plane represented by $\pi_{2}$ does not contain $P_2$, we translate it in the $\bm{z}_{3}$ direction:
\begin{equation}\label{pi2translatedPPR}
\pi'_{2} = T_{d_{3}}\pi_{2}\widetilde{T}_{d_{3}},
\end{equation}
where
\begin{equation}
T_{d_{3}} = 1 - d_{3}\dfrac{\bm{z}_{3}e_\infty}{2}.
\end{equation}
The intersection of these geometric objects is graphically depicted in figure \ref{Spheres+plane} and is computed as:
\begin{equation}\label{P2PPR}
B=\pi_{1}\vee\pi'_{2}\vee s_{1},
\end{equation}
where $B$ is a bivector. Again, this bivector represents a pair of points in the conformal geometric algebra $\mathcal{G}_{4,1}$ so:
\begin{equation}\label{P2PPRbis}
B = Q_{1}\wedge Q_2,
\end{equation}
for some null points $Q_1$ and $Q_{2}$. Clearly, if $Q_{1}$ and $Q_{2}$ belong to the workspace $\mathcal{W}$ of the robot, then there are two possible null points $P_{2}$, namely $P_{21} = Q_{1}$ and $P_{22} = Q_{2}$.  Therefore, we have two possible valid positions for the null point lying between the second and the third joint. As we will see, this means that, in this case, there are more than one set of solutions. Each one of these two null points defines a line whose inner representation, $\ell_{21}$ ($\ell_{22}$) if $P_{21}$ ($P_{22}$) is used, is computed as in \eqref{l1l2}. Similarly, the inner representation $\ell_{1}$ of another line is calculated as in \eqref{l1l2}. Now, two distinct null points $P_1$ can be obtained. Indeed, for $i=1,2$, $P_{1i}=\ell_{2i}\vee\ell_{1}$.

Finally, each pair of null points $\{P_{11},P_{21}\}$ and $\{P_{12},P_{22}\}$ allows us to compute the outer representation of an extra plane that is used to calculate the joint variable $\theta_{3}$:
\begin{equation}\label{auxiliaryplane}
	\begin{split}
		\Pi_{31} &= P_{11}\wedge P_{21}\wedge P'_{w}\wedge e_\infty,\\
		\Pi_{32} &= P_{12}\wedge P_{22}\wedge P'_{w}\wedge e_\infty.
	\end{split}
\end{equation}
Then, using the identities (\ref{distance}) and (\ref{angle}), the joint variables $d_{1},d_{2}$ and $\theta_{3}$ can be derived easily. Clearly, we are going to have two different sets of joint variables and, therefore, two different solutions:
\begin{equation}\label{jointvariablesPPR}
	\begin{split}
		d_{11} = d(P_{0},P_{11}), d_{21} = d(P_{11},P_{21}) \text{ and }\theta_{31} = \angle(\Pi_{1},\Pi_{31}),\\
		d_{12} = d(P_{0},P_{12}), d_{22} = d(P_{11},P_{22}) \text{ and }\theta_{32} = \angle(\Pi_{1},\Pi_{32}).
	\end{split}
\end{equation}
As stated in section \ref{Intro}, serial robots with 6 DoF and spherical wrist have up to eight distinct solutions for a given end-effector pose. Since the orientation problem always has two different solutions for each solution of the position problem, there is a maximum of four distinct solutions for the position problem. This maximum is reached only in two cases: (1) when the three joints that constitute the position part of the robot are revolute and (2) when the two of the three joints that constitute the position part of the robot are revolute and the other one is prismatic . Therefore, this case shows how the geometric strategy introduced in this paper gives all the solutions for the inverse kinematics and, thus, it can be considered a closed-form method. However, from now on, in order to simplify the notation for the remaining cases, we are going to consider just one of the two points that can be extracted from a given bivector and, hence, only one of the maximum of four different sets of solutions will be fully developed.

Offsets between consecutive joints are treated analogously as in the previous case. The main difference relies on the geometric object considered:
\begin{itemize}
	\item If the offset is located between the first and the second joint, then $P_2$ can be derived as in the preceding section, while $P_1$ is computed as the intersection $\ell_{1}\vee\ell'_{2}$, where $\ell'_2$ is defined as in \eqref{ell2bis}.
	\item If the offset is located between the second and the third joint, then $d_3 = \hat{d}_3 + \overline{d}_3$, where $\overline{d}_3$ is a component of the offset aligned with the $\bm{z}_3$ axis. Now, $P_{2}$ is extracted from: 
	\begin{equation}
		B = \pi_{1}\vee\pi'_{2}\vee s'_{1},
	\end{equation}
	where $s'_{1} = T_{-\overline{d}_3}T_{-a_3}s_1\widetilde{T}_{-a_3}\widetilde{T}_{-\overline{d}_3}$ is the inner representation of the sphere represented by $s_1$ after compensating the offset and $T_{-a_3},T_{-\overline{d}_3}$ are defined as in \eqref{RotorsDH}. Since the remaining geometric reasoning does not change, $P_1$ is obtained as in the case without offsets. This situation highlights another advantage of conformal geometric algebra: rotors can be applied to any geometric object which simplifies the formulation and the solution of the problem.
\end{itemize}  
Another relevant case is when the first joint is revolute, while the second and third ones are prismatic. This case is solved in a similar way, but it is interesting to point out some details. First, the null point $P_1$ is obtained as the translation of $P_0$ along $\bm{z}_1$ an amount equal to the length of the first link, i.e., $d_1$. Indeed:
\begin{equation}\label{P1RPP}
	P_{1} = T_{d_1}P_{0}\widetilde{T}_{d_1}
\end{equation}
with $T_{d_1}$ defined as in \eqref{RotorsDH}. Since the second joint is prismatic, $\theta_2$ is one of the known D-H parameters and, hence, the joint axes $\bm{z}_2$ and $\bm{z}_3$ can be computed as in \eqref{z2z3PPP}. Now, $P_2$ is derived as $P_2=\ell_2\vee \ell_3$, where $\ell_2$ and $\ell_3$ are defined as in \eqref{l3PPP} and \eqref{l1l2}. The joint variables $d_2$ and $d_3$ are determined as in (\ref{d1d2d3}). Now, we define the following two planes such as their outer representations are:
\begin{equation}
	\begin{split}
	\Pi_1 = P_0\wedge P_1\wedge P_w\wedge e_\infty,\\
	\Pi'_1 = P_0\wedge P_1\wedge P'_w\wedge e_\infty,
	\end{split}
\end{equation}
where $P'_w$ is the null vector representation of the three-dimensional point $\bm{p}'_w$ computed with the obtained values for the joint variables $d_2$ and $d_3$ and with $\theta_1 = 0$. Clearly, the first joint variable is obtained as $\theta_{1} = \angle(\Pi_1,\Pi'_1)$. 
\color{black}

\subsubsection{Two revolute joints and one prismatic (R+R+P)}\label{InPoRRP}
A scheme of the position part of a serial robot with two revolute joints and one prismatic joint is depicted in figure \ref{RRP}. As in the preceding case, there are different subcases that can be treated similarly. Because of that, only the most relevant cases are fully developed here.

\begin{figure*}[t!]
	\centering
	\includegraphics[width=10cm]{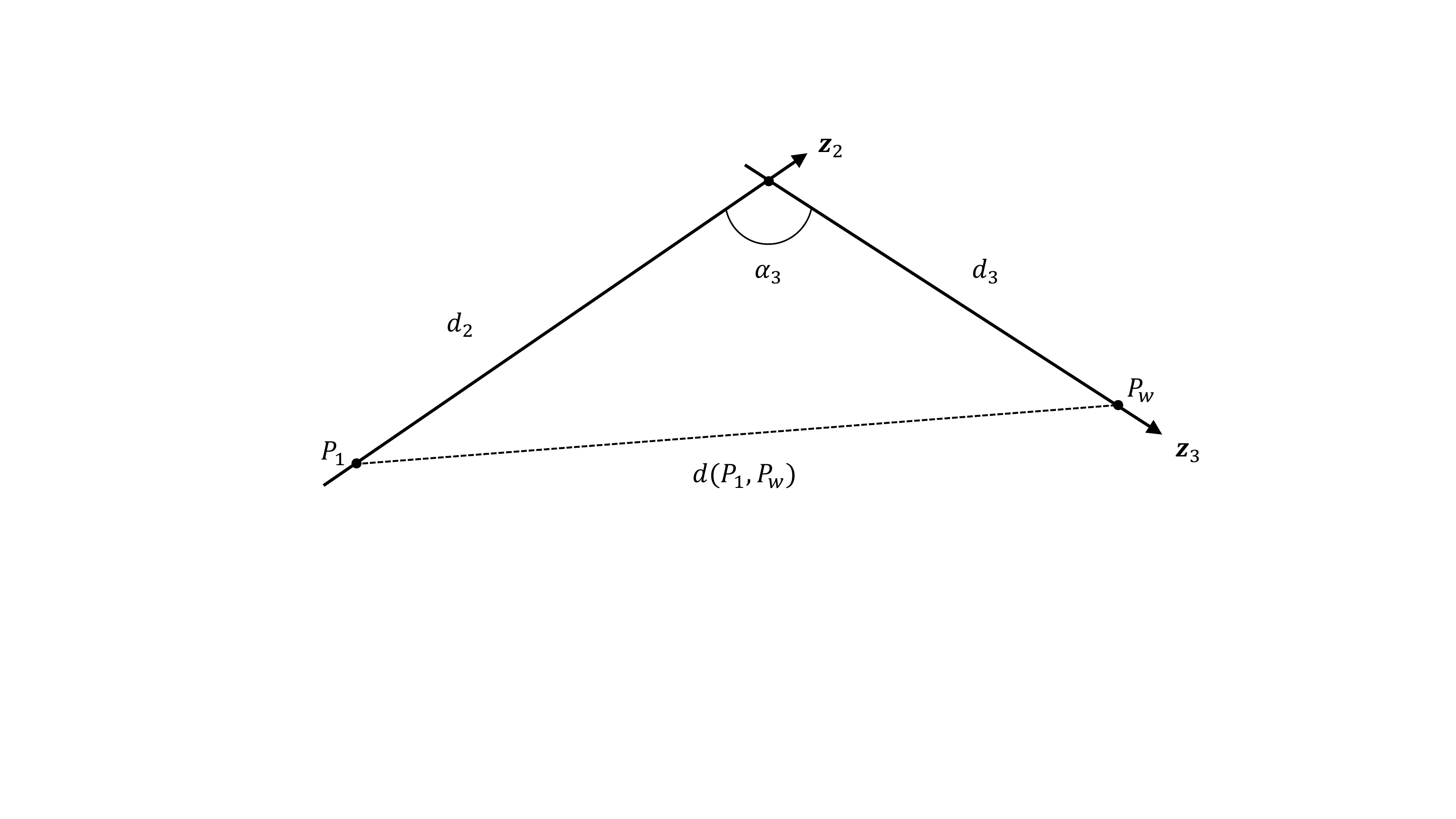}
	\caption{Triangle defined by $\bm{z}_{2}$ and $\bm{z}_{3}$.}
	\label{triangle}
\end{figure*}

First, let us consider the case where the first two joints are revolute and the third one, prismatic. In addition, for this first case, we suppose that there are no offsets between consecutive joints. The null point $P_{1}$ is computed as in \eqref{P1RPP}. Now, the remaining part of the position part of the robot is restricted to the plane determined by the joint axes $\bm{z}_{2}$ and $\bm{z}_{3}$. Furthermore, given the length of the second link, $d_{2}$, and the angle between the joint axes $\bm{z}_{2}$ and $\bm{z}_{3}$, $\alpha_{3}$, a triangle as the one depicted in figure \ref{triangle} is defined. Using the law of cosines, $d_{3}$ can be easily obtained from the quadratic equation:
\begin{equation}\label{d3}
d(P_{1},P_{w})^{2} = d_{2}^{2}+d_{3}^{2}-2d_{2}d_{3}\cos(\alpha_{3}).
\end{equation}
Now, since $d(P_{1},P_{w})>d_2\sin(\alpha_{3})$, equation (\ref{d3}) has always two distinct real solutions. Therefore, we have three different cases: (1) there are two distinct positive solutions; (2) there are two distinct negative solutions and (3) there is one positive and one negative solutions. It is clear that, since the displacements measured by the D-H parameter $d$ are always positive, the case with two distinct negative solutions is undesirable. This case only holds when
\begin{equation}\label{neg_solutions}
	d_2>d(P_1,P_w)\text{ and }d_2\cos(\alpha_{3})<0.
\end{equation}  
But, however, since $d_2\geq 0$, if $d_2\cos(\alpha_{3})<0$, then $\cos(\alpha_{3})<0$ and, thus, $\alpha_{3}\in[\pi/2,3\pi/2]$. Nevertheless, for those values of $\alpha_3$, $d_2<d(P_1,P_w)$ by definition of the D-H parameters. This is clearly the opposite of the first inequality of \eqref{neg_solutions}, which means that only the cases (1) and (3) are possible. In addition, if there is a fixed offset aligned with the $\bm{z}_{3}$ axis, then $d_{3}$ can be decomposed as $d_{3} = \widehat{d}_{3}+\overline{d}_{3}$, where, as usual, $\overline{d}_{3}$ denotes the length of this offset. It is straightforward to see that, also in this case, the joint variable $\hat{d}_{3}$ can be obtained from (\ref{d3}).

\begin{figure*}[t!]
	\centering
	\begin{subfigure}[b]{0.45\textwidth}
		\includegraphics[width=6cm]{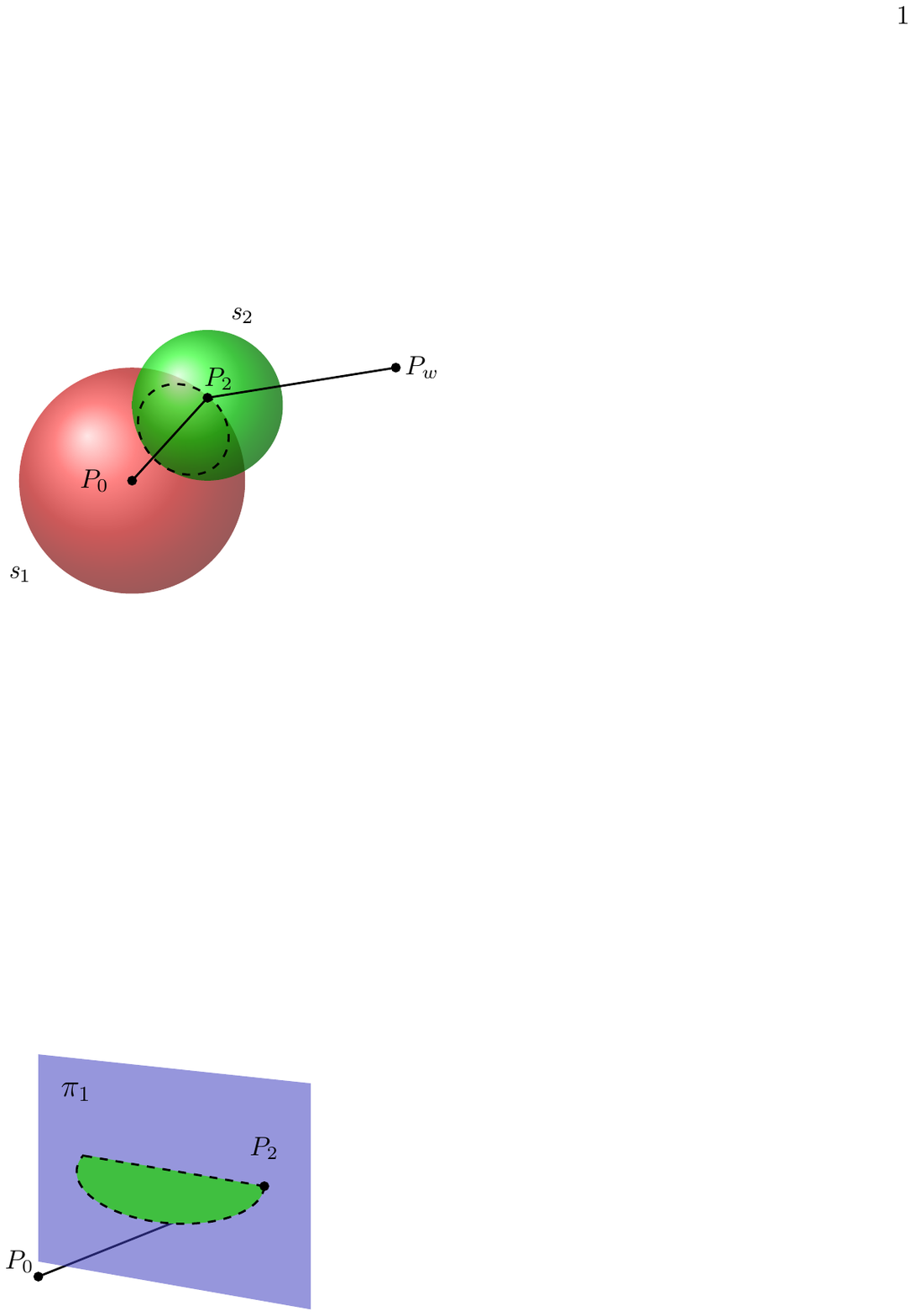}
		\caption{$P_{2}$ lying in the circle intersection of the spheres represented by $s_1$ and $s_2$.}
		\label{P2Spheres}
	\end{subfigure}
	\hspace{0.5cm}
	\begin{subfigure}[b]{0.45\textwidth}
		\includegraphics[width=6cm]{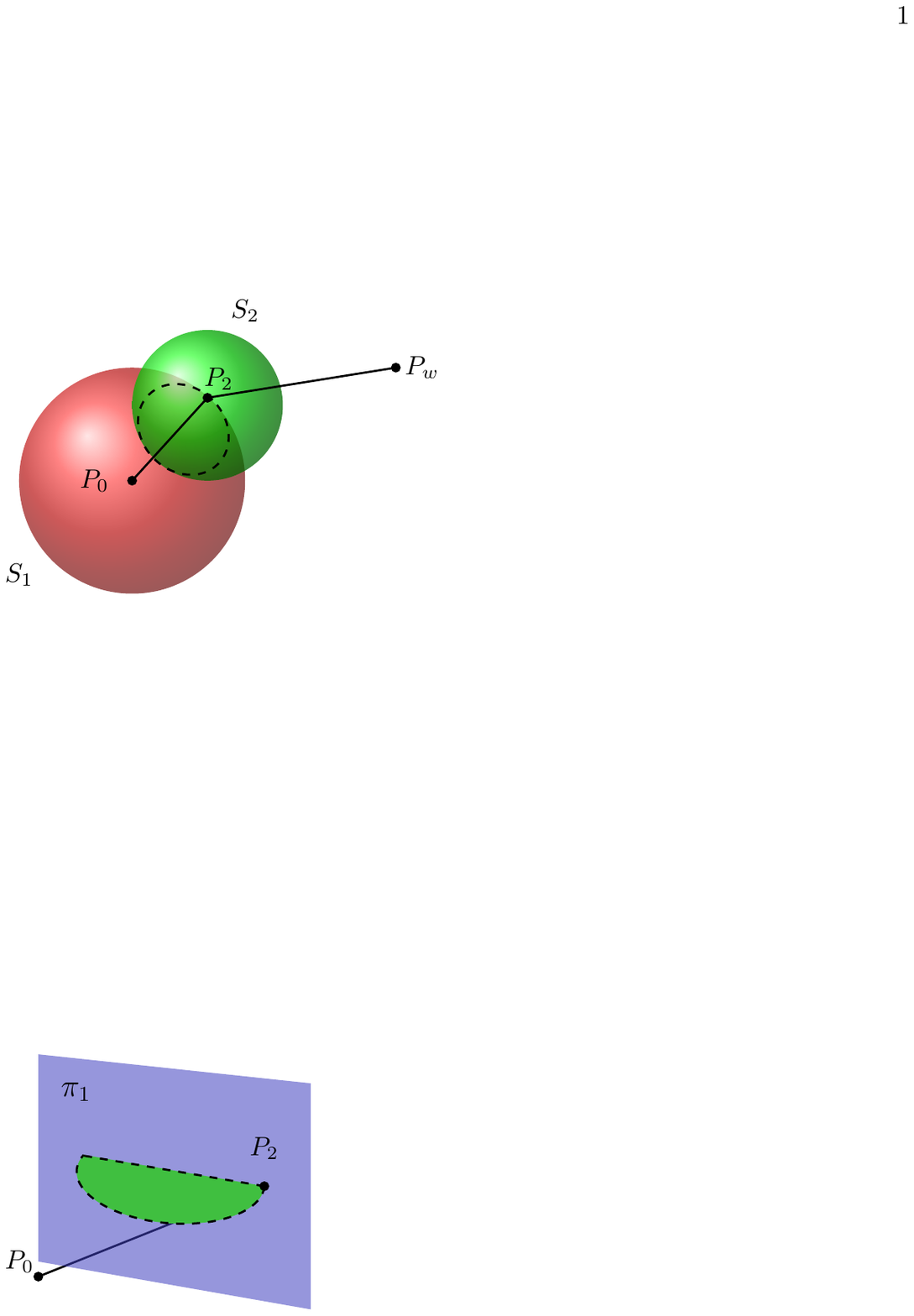}
		\caption{$P_{2}$ as the intersection of a circle and the plane represented by $\pi_1$.}
		\label{P2CirclePlane}
	\end{subfigure}
	\caption{Computation of $P_{2}$.}
\end{figure*}

Finally, in order to obtain $P_{2}$, we define two distinct spheres and one plane through their inner and outer representations:
\begin{equation}\label{S1S2p1RRP}
\begin{split}
s_{1} &= P_{1} - \dfrac{1}{2}d_{2}^{2}e_\infty,\\
s_{2} &= P_{w} - \dfrac{1}{2}d_{3}^{2}e_\infty,\\
\Pi_{1} &= P_{0}\wedge P_{1}\wedge P_{w}\wedge e_\infty.
\end{split}
\end{equation}
Then, the intersection of these geometric objects is computed (and shown graphically in figures \ref{P2Spheres} and \ref{P2CirclePlane}):
\begin{equation}\label{P2RRP}
B = s_{1}\vee s_{2}\vee\pi_{1},
\end{equation}
where two different null points $P_{2}$ can be extracted from $B$. Since we have already computed $d_{3}$, we use the null points $P_{0},P_{1}$ and $P_{2}$ to derive the remaining joint variables, namely $\theta_{1}$ and $\theta_{2}$. The following auxiliary geometric objects are defined for that purpose:
\begin{itemize}
	\item If the revolute joints have parallel joint axes:
	\begin{equation}
	L_{1} = P_{1}\wedge P_{2}\wedge e_\infty\text{ and } L_{2} = P_{2}\wedge P_{w}\wedge e_\infty.
	\end{equation}
	\item If the revolute joints have not parallel joint axes:
	\begin{equation}
	\Pi_{2} = P_{0}\wedge P_{1}\wedge P_2\wedge e_\infty\text{ and }\Pi_{3} = P_{1}\wedge P_{2}\wedge P_{w}\wedge e_\infty.
	\end{equation}
\end{itemize}
\begin{figure*}[t!]
	\centering
	\includegraphics[width=10cm]{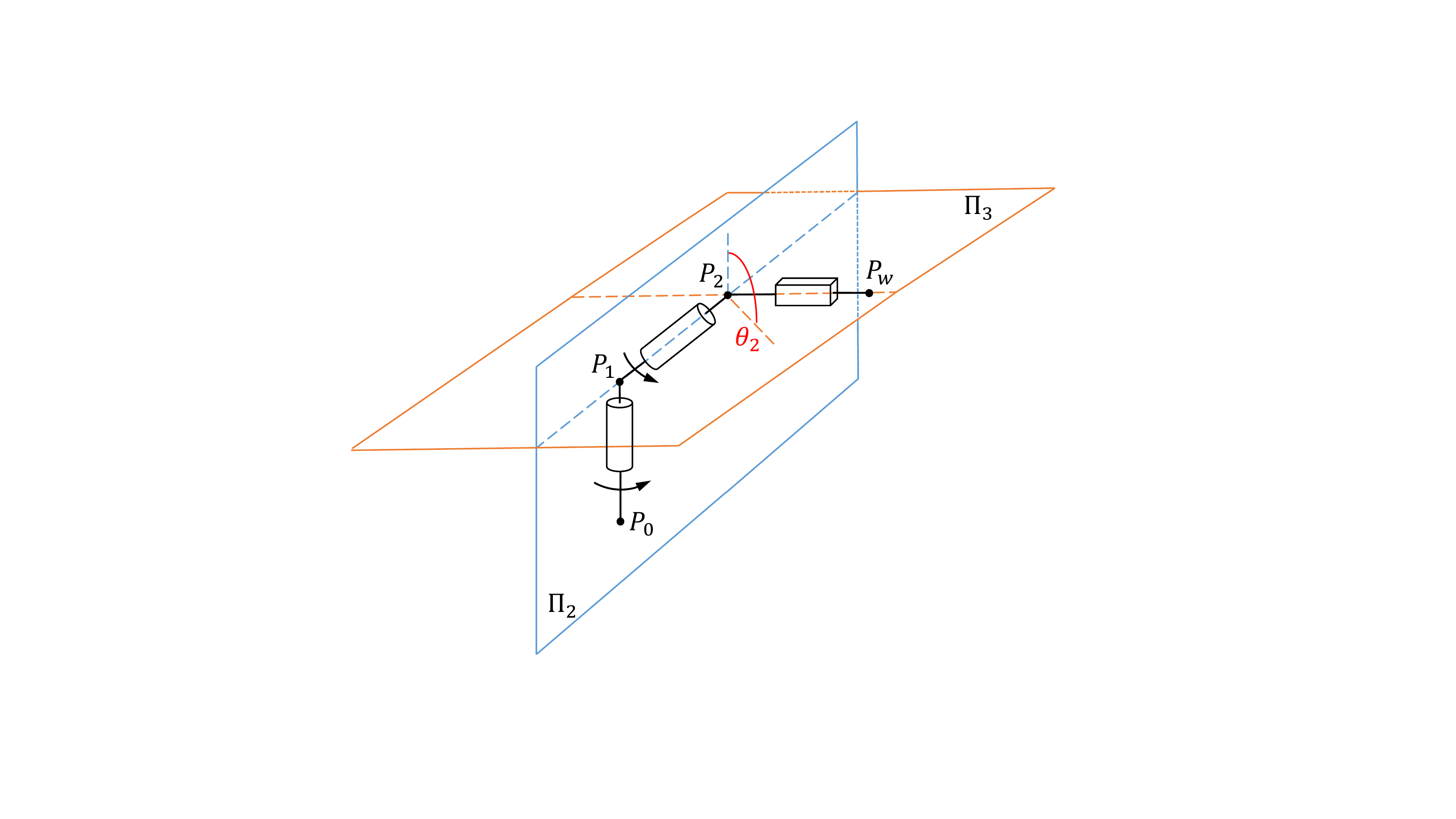}
	\caption{Computation of $\theta_2$ as the angle between $\Pi_{2}$ and $\Pi_3$ for the case of non-parallel revolute joint axes.}
	\label{RRR_method1}
\end{figure*}
Now, for the parallel case, the joint variable $\theta_2$ is computed as follows:
\begin{equation}
\theta_{2} =\angle(L_{1},L_{2}).
\end{equation}
Since the joint variables $\theta_2$ and $d_3$ are already known, the position of the end-effector with $\theta_1 = 0$ can be computed. If we denoted such a three-dimensional point $\bm{p}'_w$, we can define an extra plane whose outer representations is:
\begin{equation}
	\Pi'_1 = P_0\wedge P_1\wedge P'_w\wedge e_\infty,
\end{equation} 
where $P'_w$ is the null vector representation of $\bm{p}'_w$. Now, $\theta_{1} =\angle(\Pi_{1},\Pi'_{1})$. For the other case (shown graphically in figure \ref{RRR_method1}), we have that:
\begin{equation}
\theta_{2} =\angle(\Pi_{2},\Pi_{3}),
\end{equation}
where, again, $\theta_1$ is computed as $\theta_{1} =\angle(\Pi_{1},\Pi'_{1})$, where $\Pi'_1$ is defined with the already obtained values of $\theta_2$ and $d_3$ and with $\theta_1= 0$.\color{black}

Offsets are treated analogously to the preceding cases. Two different situations arise:
\begin{itemize}
	\item If the offset is placed between the second and the third joints, then we translate the null point $P_w$ as follows:
	\begin{equation}
		P'_w = T_{-\overline{d}_3}T_{-a_3}P_w\widetilde{T}_{-a_3}\widetilde{T}_{-\overline{d}_3},
	\end{equation}
	where $T_{-\overline{d}_3}$ and $T_{-a_3}$ are defined as in \eqref{RotorsDH}. Now, the rest of the reasoning is as in the case without offsets. The null point $P_1$ is calculated as in \eqref{P1RPP} and, hence, a triangle such as the one depicted in figure \ref{triangle} can be defined using, instead of $P_w$, $P'_w$.
	\item If the offset is placed between the first and the second joints, we can simply translate the null point $P_0$ in the direction $\bm{x}_1$ and amount equal to $a_2$, i.e., to compute $P'_0 = T_{a_2}P_0\widetilde{T}_{a_2}$, where $T_{a_2}$ is defined as in \eqref{RotorsDH}. The remaining steps are as in the case without offsets, using $P'_0$ instead of $P_0$.
\end{itemize}
Finally, let us consider another subcase: the first joint is revolute, the second joint is prismatic and the third joint is again revolute. The null points $P_{0}$ and $P_{1}$ are obtained as in the preceding case. To calculate the null point $P_{2}$, the geometric reasoning used for the case where there are two prismatic joints followed by one revolute is applied. Therefore, the planes and the sphere represented by $\pi_{1},\pi_{2},\pi_{3}$ and $s_{1}$ are defined as in (\ref{plano1PPP}),(\ref{pi2PPR}),(\ref{S1PPR}) and (\ref{auxiliaryplane}). Finally, the joint variables are obtained as in the above-mentioned case.

\subsubsection{Three revolute joints (R+R+R)}\label{InPoRRR}
A scheme of the position part of a serial robot with three revolute joints is depicted in figure \ref{RRR}. First, the case without offsets is developed.

As in the preceding case, the null point $P_{1}$ is computed as $P_{1} = T_{d_{1}}P_{0}\widetilde{T}_{d_{1}}$, where $T_{d_1}$ is defined as in \eqref{RotorsDH}. Once $P_{1}$ has been obtained, the null point $P_{2}$ is computed by intersecting two spheres and one plane whose outer and inner representations are:
\begin{equation}\label{S1}
\begin{split}
\Pi_{1} &= P_{0}\wedge P_{1}\wedge P_{w}\wedge e_\infty,\\
s_{1} &= P_{1} - \dfrac{1}{2}a_{2}^{2}e_\infty,\\
s_{2} &= P_{w} - \dfrac{1}{2}(a_{3}+d_3)^{2}e_\infty.
\end{split}
\end{equation}
The intersection of these three geometric objects is, again, a bivector:
\begin{equation}\label{B2}
B = s_{1}\vee s_{2}\vee\pi_{1},
\end{equation}
where, as in the preceding cases, two different points $P_{2}$ can be extracted from $B$, each one leading to a different solution.

Now, two distinct cases are considered to obtain the joint variables: two joint axes are parallel or no joint axes are parallel. The case where the three joint axes are parallel is not considered because it will mean that the position problem is a two-dimensional problem instead of a three-dimensional one, i.e., that the robot has less than 6 DoF. Hence:
\begin{figure*}[t!]
	\centering
	\includegraphics[width=11cm]{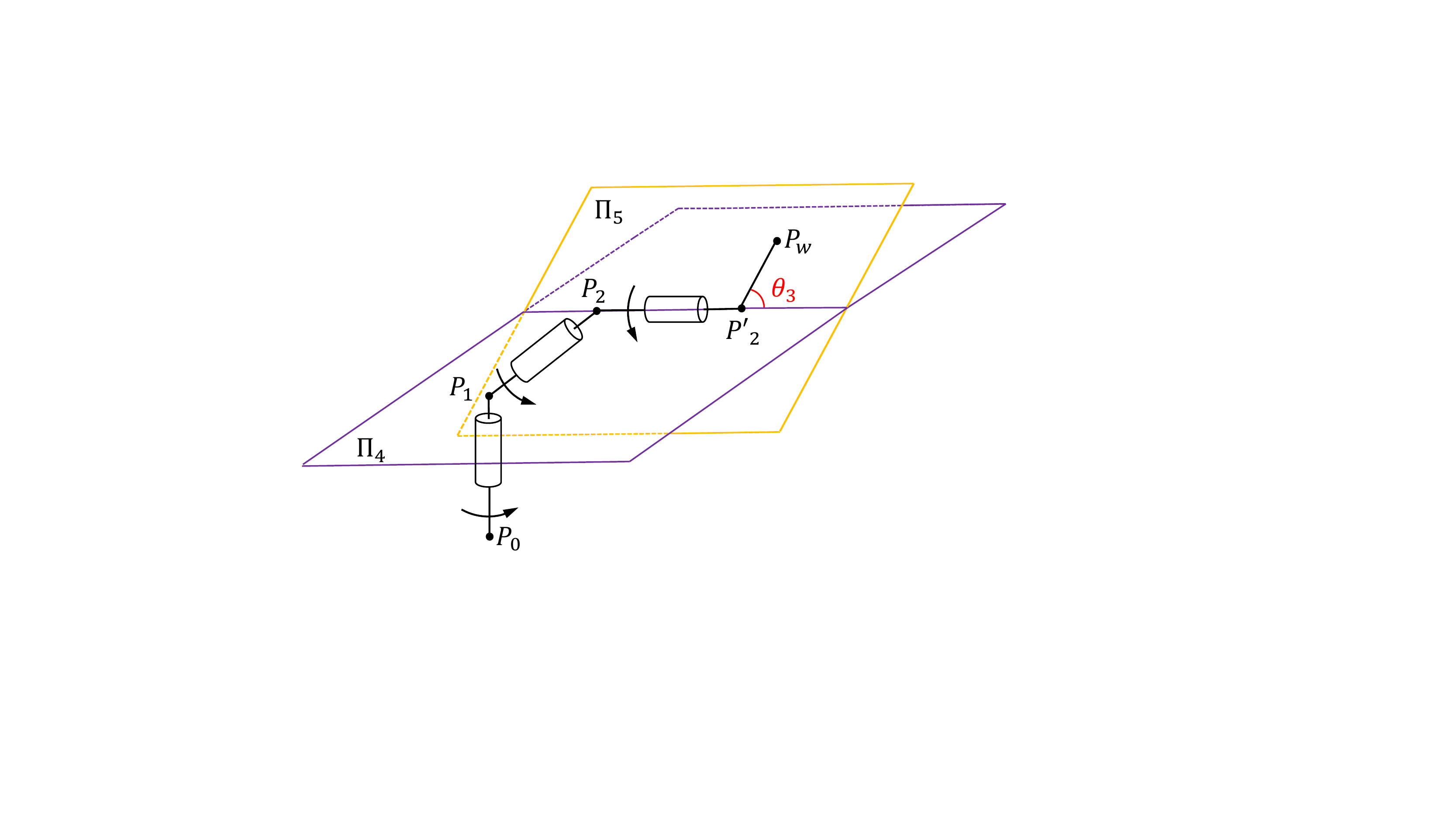}
	\caption{Computation of $\theta_3$ as the angle between $\Pi_4$ and $\Pi_5$ for the case of non-parallel revolute joint axes.}
	\label{RRR_method2}
\end{figure*}
\begin{itemize}
	\item If there are two joints whose joint axes are parallel, then let us suppose that the two parallel joint axes are the last two (the other cases are analogous). Then, the outer representation of the geometric objects required for computing the joint variables are defined as:
	\begin{equation}\label{l1l2l3}
		\begin{split}
			L_{1} &= P_{0}\wedge P_{1}\wedge e_\infty,\\
			L_{2} &= P_{1}\wedge P_{2}\wedge e_\infty,\\
			L_{3} &= P_{2}\wedge P_{w}\wedge e_\infty,
		\end{split}
	\end{equation}
	and the joint variables are:
	\begin{equation}
		\theta_{1} =\angle(\bm{x}_{1},\Pi_{1}), \theta_{2} =\angle(L_{1},L_{2})\text{ and }\theta_{3} =\angle(L_{2},L_{3}),
	\end{equation}
	where $\angle(\bm{x}_{1},\Pi_{1})$ cannot be computed using equation \eqref{angle} since it is only defined for blades of the same grade. However, the following relation holds:
	\begin{equation}\label{angle_plane_vector}
		\angle(\bm{x}_{1},\Pi_{1}) =\dfrac{\pi}{2} - \angle(\bm{n},\bm{x}_1),
	\end{equation} 
	where $\bm{n}$ is the vector normal to $\Pi_1$ (and can be easily retrieved from its inner representation $\pi_1$), so the computation of $\angle(\bm{x}_{1},\Pi_{1})$ reduces to the computation of $\angle(\bm{x}_{1},\bm{n})$. \color{black}Interestingly, this coincides with the approach followed in \cite{HildenbrandZamoraCorrochano08,ZamoraCorrochano04}.
	\item If there are no revolute joints with parallel joint axes, we have that:
	\begin{equation}
		\begin{split}
			\Pi_{2} &= P_{0}\wedge P_{1}\wedge P_{2}\wedge e_\infty,\\
			\Pi_{3} &= P_{1}\wedge P_{2}\wedge P_{w}\wedge e_\infty,
		\end{split}
	\end{equation}
	and, therefore:
	\begin{equation}
		\theta_{2} =\angle(\Pi_{2},\Pi_{3}).
	\end{equation}
	Then, once $\theta_2$ is computed, the joint axis of the second joint, $\bm{z}_2$, can be computed from the three-dimensional points $\bm{p}_1$ and $\bm{p}_2$ (which are the Euclidean points whose null vector representation are $P_1$ and $P_2$ respectively). \color{black}Now, $\bm{z}_3$ is calculated as in \eqref{z2z3PPP} and we translate the null vector $P_2$ along $\bm{z}_3$ an amount equal to $a_3$:
	\begin{equation}
		P'_2 = T_{a_3}P_{2}\widetilde{T}_{a_3}.
	\end{equation}
	This extra null vector allows us to define two additional planes:
	\begin{equation}
		\begin{split}
			\Pi_4 &= P_1\wedge P_2\wedge P'_2\wedge e_\infty,\\
			\Pi_5 &= P_2\wedge P'_2\wedge P_w\wedge e_\infty,
		\end{split}
	\end{equation}
	which, in turn, allow us to compute $\theta_3$ as follows:
	\begin{equation}
		\theta_{3} =\angle(\Pi_{4},\Pi_{5}).
	\end{equation}
	This is graphically depicted in figure \ref{RRR_method2}. Finally, $\theta_1$ is computed as in the previous cases.
\end{itemize}
Finally, the offsets are treated by transforming the proper geometric object. For instance, if the offset is placed between the first and the second joint, then $P_{0}$ is transformed into $P'_{0} = T_{a_2}P_0\widetilde{T}_{a_2}$, where $T_{a_2}$ is defined as in \eqref{RotorsDH}. Then, the remaining steps are as in the case without offsets using $P'_0$ instead of $P_0$. If, conversely, the offset is placed between the second and the third joint, then the sphere represented by $s_2$ is transformed into $s'_2 = T_{-a_3}s_2\widetilde{T}_{-a_3}$, where, again, $T_{-a_3}$ is computed as in \eqref{RotorsDH}. Therefore, the null point $P_{2}$ is extracted from the intersection bivector $\pi_{1}\vee s_{1}\vee s'_{2}$ and the remaining steps are the same as the ones in the case without offsets.

\subsection{Solution of the orientation problem}\label{SOP}
\begin{figure*}
	\begin{center}
		\includegraphics[width=6cm]{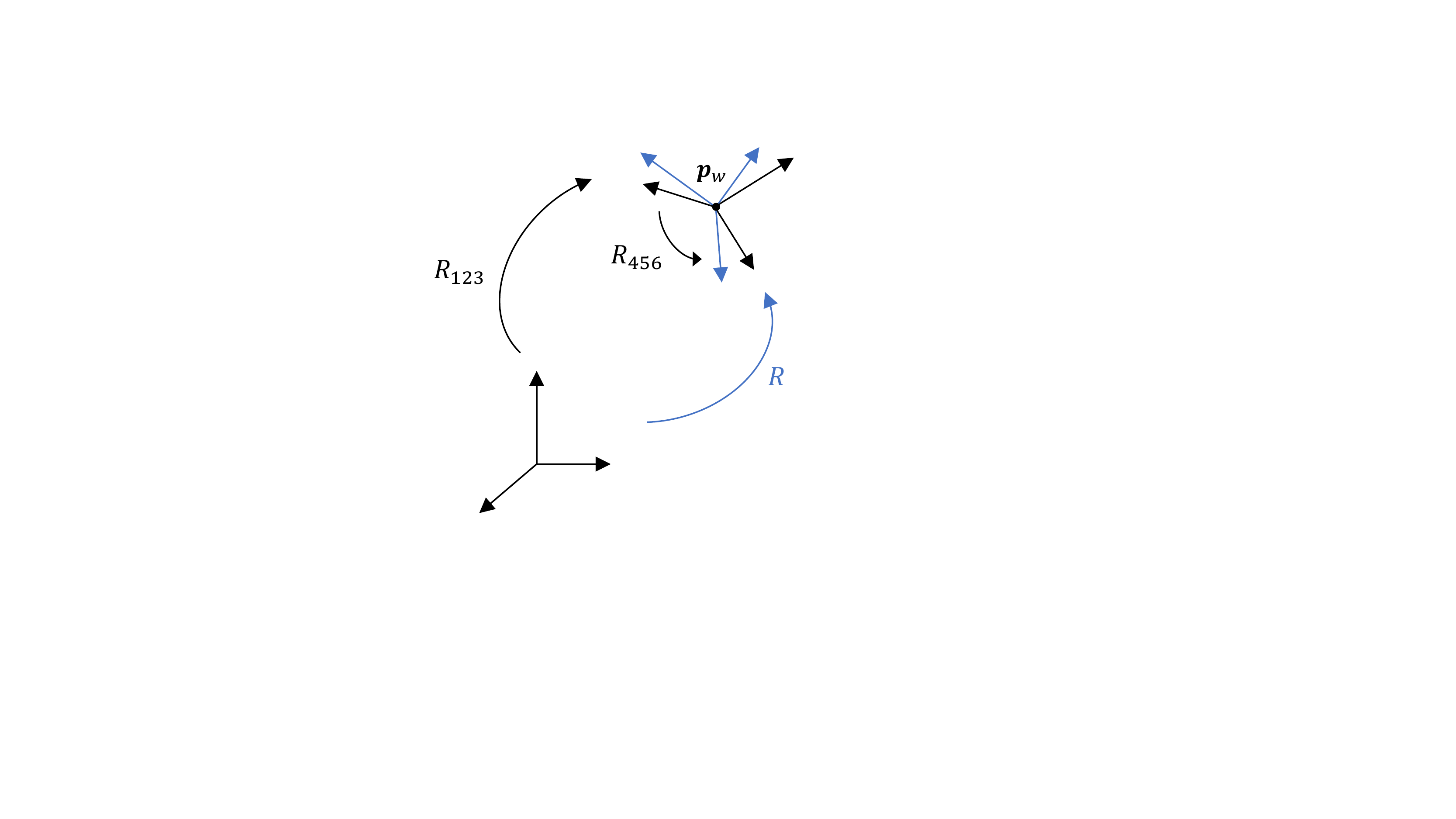}
		\caption{Relation between the orientations}
		\label{Rotaciones}
	\end{center}
\end{figure*}

Once the position problem is solved, the value of the joint variables $q_{1},q_{2},q_{3}$ is known and it only remains to find $q_{4},q_{5}$ and $q_{6}$. With $q_{1},q_{2},q_{3}$, we compute the rotor defining the orientation of the frame attached to $\bm{p}_{w}$ under the effect of these joints and we denote it by $R_{123}$. Now, recall that $R_{I_3,R}$ denotes the rotor representing the target orientation of the end-effector. Then, the rotor that defines the rotation between $R_{123}$ and $R_{I_3,R}$ can be obtained from the relation (figure \ref{Rotaciones}):
\begin{equation}\label{desiredRotor}
R_{I_3,R} = R_{123}R_{456},
\end{equation}
where the rotor $R_{456}$ only depends on the joint variables $q_{4},q_{5}$ and $q_{6}$. Since these joints are revolute, $q_{4},q_{5}$ and $q_{6}$ correspond to $\theta_{4},\theta_{5}$ and $\theta_{6}$, respectively.

Now, we split $R_{456}$ into three different rotors as follows:
\begin{equation}\label{RotorSplit}
R_{456}=R_{4}R_{5}R_{6},
\end{equation}
where:
\begin{equation}
	R_{i} = \cos\left(\dfrac{\theta_{i}}{2}\right)-\sin\left(\dfrac{\theta_{i}}{2}\right)B_{i},
\end{equation}
where $\theta_{i}$ is the angle of rotation $B_{i}$ is the bivector defining the rotation plane. Clearly, the angle extracted from each one of these rotors will correspond to one of the sought joint variables. We expand (\ref{RotorSplit}) and obtain:
\begin{equation}\label{BivectorSenCos}
\begin{split}
& R_{456} = R_{4}R_{5}R_{6}=\text{c}_4\text{c}_5\text{c}_6-\text{c}_4\text{c}_6\text{s}_5B_{5}-\text{c}_5\text{c}_6\text{s}_4B_{4}-\text{c}_4\text{c}_5\text{s}_6B_{6}+\\
&+\text{c}_6\text{s}_4\text{s}_5B_{4}B_{5}+\text{c}_4\text{s}_5\text{s}_6B_{5}B_{6}+\text{c}_5\text{s}_4\text{s}_6B_{4}B_{6}-\text{s}_4\text{s}_5\text{s}_6B_{4}B_{5}B_{6},
\end{split}
\end{equation}
where $\text{c}_i = \cos(\theta_{i}/2)$ and $\text{s}_i = \sin(\theta_{i}/2)$. In addition, $R_{456}$ can also be written:
\begin{equation}\label{rotoreuler}
R_{456} = \cos\biggl(\dfrac{\theta}{2}\biggr) - \sin\biggl(\dfrac{\theta}{2}\biggr)B_{456}
\end{equation}
for a known angle $\theta$ and bivector $B_{456}$. But, since $R_{I_3,R}$ and $R_{123}$ are rotors in $\mathcal{G}_3$ (by definition), then $R_{456}$ is also a rotor in $\mathcal{G}_3$ and, thus, we can write $B_{456}$ as a linear combination with respect to the basis bivectors of $\mathcal{G}_{3}$:
\begin{equation}\label{BivectorBasis}
B_{456} = \beta_{1}e_{23}+\beta_{2}e_{13}+\beta_{3}e_{12}
\end{equation}
for some $\beta_{1},\beta_{2},\beta_{3}\in\mathbb{R}$. Then, we can rewrite $R_{456}$ as:
\begin{equation}\label{rotoreulerbivector}
R_{456} = \cos\biggl(\dfrac{\theta}{2}\biggr) - \sin\biggl(\dfrac{\theta}{2}\biggr)\beta_{1}e_{23}-\sin\biggl(\dfrac{\theta}{2}\biggr)\beta_{2}e_{13}-\sin\biggl(\dfrac{\theta}{2}\biggr)\beta_{3}e_{12}.
\end{equation}
Now, as happens with Euler angles, different conventions can be adopted. Depending on the convention used, a particular set of equations is obtained. For instance, by setting
\begin{equation}\label{EulerConvention1}
B_{4}=e_{12}, B_{5}=e_{13}\text{ and } B_{6}=e_{12},
\end{equation}
that corresponds to the Euler angles convention $ZYZ$, the following relations hold:
\begin{equation}
B_{4}B_{5} = -e_{23}, B_{5}B_{6} = e_{23}, B_{4}B_{6} = -1\text{ and } B_{4}B_{5}B_{6} = e_{13}.
\end{equation}
Hence, by regrouping the terms of (\ref{BivectorSenCos}) and equating them to (\ref{rotoreulerbivector}), the following set of equations is obtained:
\begin{equation}\label{EulerConventionAlpha1}
\left.\begin{split}
\alpha &= \text{c}_4\text{c}_5\text{c}_6 - \text{c}_5\text{s}_4\text{s}_6\\
\beta'_{1} &= \text{c}_4\text{s}_5\text{s}_6 - \text{c}_6\text{s}_4\text{s}_5\\
\beta'_{2} &= -\text{c}_4\text{c}_6\text{s}_5 - \text{s}_4\text{s}_5\text{s}_6\\
\beta'_{3} &= -\text{c}_5\text{c}_6\text{s}_4 - \text{c}_4\text{c}_5\text{s}_6\\
\end{split}\right\}
\end{equation}
where, again, $\text{c}_i = \cos(\theta_{i}/2)$ and $\text{s}_i = \sin(\theta_{i}/2)$, while $\alpha =\cos(\theta/2)$ and $\beta'_{i} = -\sin(\theta/2)\beta_{i}$. Now, by squaring and adding the first and last identities, we obtain:
\begin{equation}
\begin{split}
\alpha^{2}+(\beta'_{3})^{2} =\cos^{2}\biggl(\dfrac{\theta_{5}}{2}\biggr),
\end{split} 
\end{equation}
and, thus,
\begin{equation}\label{theta5}
\theta_{5} = \dfrac{\cos^{-1}\biggl(\pm\sqrt{\alpha^{2}+(\beta'_{3})^{2}}\biggr)}{2}.
\end{equation}
Finally, if $\cos(\theta_{5}/2)\neq 0$ and $\sin(\theta_{5}/2)\neq 0$, then:
\begin{equation}
\alpha' = \dfrac{\alpha}{\cos(\theta_{5}/2)}\,\text{ and }\,\beta''_{2} = \dfrac{-\beta'_{2}}{\sin(\theta_{5}/2)}
\end{equation}
and, therefore:
\begin{equation}\label{theta4y6}
\begin{split}
\theta_{6} = \sin^{-1}(\alpha')+\sin^{-1}(\beta''_{2})\text{ and }\theta_{4} = \sin^{-1}(\alpha')-\sin^{-1}(\beta''_{2}),
\end{split}
\end{equation}
which completes the resolution of the orientation problem. In case $\theta_5 = 0$ or $\pi$, we are in the so-called \textit{representation singularity} of the Euler angles convention $ZYZ$ \cite{Siciliano08a} and, hence, we can only obtain the value of the sum $\theta_4+\theta_6$. Indeed, from \eqref{BivectorSenCos} we can easily deduce that $\theta_4+\theta_6 = 2\cos^{-1}(\alpha)$ if $\theta_5 = 0$ and $\theta_4+\theta_6 = 2\cos^{-1}(-\alpha)$ if $\theta_5 = \pi$.

Although the obtained solutions are equivalent to Euler angles, the advantages of the proposed method are evident. Rotor manipulation is more geometrically intuitive than manipulating Euler angles. In fact, here, rotor manipulation reduces to bivector manipulation, which can be seen as a "linearization" of the problem. In addition, it avoids the use of rotation matrices. Although rotation matrices have good properties that facilitate their computations, they are still more challenging to manipulate than rotors. In addition, their geometric interpretation is not intuitive, while, as stated before, this is the case for rotors.

\section{Solutions for redundant robots with spherical wrist}\label{RRSW}
As stated in section \ref{Intro}, there is an infinite number of solutions for the inverse kinematics of redundant serial robots. In \cite{ZaplanaBasanez17}, redundant robots are reduced to non-redundant ones by the parametrization of the so-called \textit{redundant joints}. Once the analytical solutions for the inverse kinematics have been obtained, particular instances for the parametrized joints are given. Hence, for each instance, a set of a maximum of sixteen solutions is obtained (recall that, for non-redundant robots, this is the upper bound for the number of distinct solutions associated with a given end-effector pose). This can be regarded as the addition of extra conditions in an undetermined problem. Therefore, these particular solutions form a subset of the set of all possible solutions for a given target pose of the end-effector. 

Following this idea, the geometric strategies introduced and developed in section \ref{SPP} are extended to serial robots with 7 DoF and spherical wrist. Here, the extra information is given in the form of extra null points $P_{i}$ that allow the definition of the geometric entities involved in the resolution. As it will be seen in the following cases, the addition of these extra points entails the evaluation of one of the joint variables. In addition, if the identification of the redundant joints developed in \cite{ZaplanaBasanez17} is applied to these manipulators, it is possible to know exactly which null points $P_{i}$ are needed in order to completely solve the position problem.

For each one of the cases studied in the preceding section, an extra degree of freedom -- that corresponds either to a prismatic or revolute joint -- is added. Several distinct combinations arise, but as in section \ref{SPP}, only the most relevant are fully developed in order not to repeat the already used geometric reasonings. In particular, the cases to be considered are: P+P+P+P, P+P+R+P, R+R+P+R and R+R+R+R where, again, P denotes a prismatic joint and R denotes a revolute joint.

\subsection{P+P+P+P}
The objective is to find $P_{1},P_{2}$ and $P_{3}$ to obtain the joint variables $d_{1},d_{2},d_{3}$ and $d_{4}$. Since all the joints are prismatic, the joint axes $\bm{z}_{2},\bm{z}_{3}$ and $\bm{z}_{4}$ can be calculated as in (\ref{z2z3PPP}). Now, the inner representation of two distinct planes:
\begin{equation}\label{pi1pi2PPPP}
\begin{split}
\pi_{1} &= \bm{z}_{1}\times\bm{z}_{2},\\
\pi_{2} &= \bm{z}_{3}\times\bm{z}_{4} + d(P_{0},P_{w})e_\infty
\end{split}
\end{equation}
and two distinct lines is defined:
\begin{equation}\label{l1l4PPPP}
\begin{split}
\ell_{1} &=\bm{z}_{1}e_{123}- (e_0\wedge\bm{z}_{1})e_{123}e_\infty,\\
\ell_{2} &=\bm{z}_{4}e_{123} - (\bm{p}_{w}\wedge\bm{z}_{4})e_{123}e_\infty.
\end{split}
\end{equation}
The intersection of the planes represented by $\pi_{1}$ and $\pi_{2}$ is a line containing the null point $P_{2}$. Indeed, its outer representation is $L_{2} = \pi_{1}\vee\pi_{2}$. Now, to determine uniquely $P_{2}$, another plane is needed. Such a plane contains the joint axes $\bm{z}_{2}$ and $\bm{z}_{3}$. However, it also requires an extra point for setting its distance to $P_{0}$. Let us define $P_{1}$ as the null point that verifies:
\begin{equation}
P_{1}\wedge L_{1} = 0\quad\text{and}\quad d(P_{0},P_{1})=d_{1}.
\end{equation}
Notice that the definition of $P_{1}$ is equivalent to setting $d_{1}$ at a particular value. Now, the inner representation of the new plane can be defined as:
\begin{equation}
\pi_{3} = \bm{z}_{2}\times\bm{z}_{3} + d_{1}e_\infty
\end{equation} 
and, therefore, $P_{2}$ can be extracted from the intersection bivector $\pi_{3}\vee\ell_{2}$.

Once $P_{0}, P_{1}$ and $P_{2}$ have been obtained, $P_{3}$ is found as the intersection of the lines represented by $\ell_{3}$ and $\ell_{4}$, where:
\begin{equation}
\ell_{3} =\bm{z}_{3}e_{123} - (\bm{p}_{2}\wedge\bm{z}_{3})e_{123}e_\infty.
\end{equation}
Again, $\bm{p}_{2}$ represents the three dimensional vector whose null vector is $P_{2}$ and that is recovered using the projection $P:\mathbb{R}^{4,1}\to\mathbb{R}^3$. Finally, the remaining joint variables $d_{2},d_{3}$ and $d_{4}$ are calculated as in (\ref{d1d2d3}).

\subsection{P+P+R+P}
Again, the joint axes $\bm{z}_{2}$ and $\bm{z}_{3}$ are calculated as in (\ref{z2z3PPP}). However, since the third joint is revolute, $R_{\theta_{3}}$ is not a constant rotor and, thus, the joint axis $\bm{z}_{4}$ cannot be computed. The inner representations of the line $\ell_{1}$ and the plane $\pi_{1}$ are defined as in (\ref{pi1pi2PPPP}) and (\ref{l1l4PPPP}). Now, an extra point is needed to completely describe the position of $\bm{z}_{4}$. Let us denote by $P_{3}$ the null point verifying:
\begin{equation}
\angle(\Pi_{1},L_{3})=\theta_{3},
\end{equation}
where $\ell_{3}=\bm{z}_{3}e_{123} - (\bm{p}_{3}\wedge\bm{z}_{3})e_{123}e_\infty$. Since they are represented by blades of different grade, the angle between $\Pi_1$ and $L_3$ cannot be computed using equation \eqref{angle}. However, we can define a line whose outer representation $L$ satisfies:
\begin{equation}
	\angle(\Pi_{1},L_{3}) = \dfrac{\pi}{2} - \angle(L,L_3).
\end{equation}
Such a line is constructed as follows. The inner representation of $\Pi_1$ is $\pi_1 = \bm{n} + \delta e_\infty$, where $\bm{n}$ denotes the vector normal to the plane. Since $\Pi_1$ passes through the origin, its inner representation reduces to $\pi_1 = \bm{n}$. Therefore, we can define a line passing also through the origin whose inner representation is $\ell = \bm{n}e_{123}$. Finally, its outer representation, $L = I_5\ell$, gives us the desired line. Furthermore, this definition of the null point $P_{3}$ is equivalent to setting the joint variable $\theta_3$ to a particular value. Now, $P_{3}$ is translated along the $\bm{z}_{3}$ axis:
\begin{equation}
P_{2} = T_{-d_3}P_{3}\widetilde{T}_{-d_3}
\end{equation}
where $T_{-d_3}$ is defined as in \eqref{RotorsDH}. Finally, we define the inner representation $\ell_2$ of a line as in \eqref{l1l2}. The null point $P_{1}$ is found as the intersection point $\ell_{1}\vee\ell_{2}$. Once the points $P_{0},P_{1},P_{2}$ and $P_{3}$ have been obtained, the joint variables $d_{1},d_{2}$ and $d_{4}$ can be calculated as in equation (\ref{d1d2d3}).

\subsection{R+R+P+R}
The null point $P_{0}$ is translated along the joint axis $\bm{z}_{1}$ by an amount equal to the length of the first link, $d_1$. Indeed, $P_1 = T_{d_1}P_0\widetilde{T}_{d_1}$, where $T_{d_1}$ is defined as in \eqref{RotorsDH}. Now, the inner representation of a sphere is defined as:
\begin{equation}\label{S1RRPR}
s_{1} = P_{w}-\dfrac{1}{2}d_4^{2}e_\infty,
\end{equation}
where $d_4$ corresponds to the length of the fourth link. To continue, an extra point is needed. Let us denote by $P_{3}$ the  null point that satisfies:
\begin{equation}\label{P3}
P_{3}\wedge S_{1}=0\quad\text{and}\quad d(P_{1},P_{3}) = d.
\end{equation}
Since the second and third joint axes belong to the same plane, their links can be regarded as the sides of a triangle where, in addition, the third side is the line segment defined by $P_{1}$ and $P_{3}$. As shown in section \ref{InPoRRP}, the law of cosines can be applied in this situation to deduce the expression for the joint variable $d_{3}$. The length of both sides, as well as the angle between the joint axes, are known: $d_{2},d$ and $\alpha_{3}$. Therefore:
\begin{equation}\label{cosines_RRPR}
d^{2} = d_{2}^{2}+d_{3}^{2} - 2d_{2}d_{3}\cos(\alpha_{3}),
\end{equation}
where, as explained in section \ref{InPoRRP}, there are either two distinct positive solutions or a unique positive solution for $d_{3}$. This proves the equivalence between the extra point $P_{3}$ and the joint variable $d_{3}$, i.e., given a particular instance of $d_3$, we use equation \eqref{cosines_RRPR} to compute $d$ and, with $d$, we define the null point $P_3$ as in \eqref{P3}. Finally, the null point $P_{2}$ is extracted from the intersection bivector defined by the following geometric objects:
\begin{equation}
\begin{split}
s_{2} &= P_{1}-\dfrac{1}{2}d_{2}^{2}e_\infty,\\
s_{3} &= P_{3}-\dfrac{1}{2}d_{3}^{2}e_\infty,\\
\Pi_{1} &= P_{0}\wedge P_{1}\wedge P_{3}\wedge e_\infty.
\end{split}
\end{equation}
Once all the points have been obtained, it is easy to recover the joint variables $\theta_{1},\theta_{2}$ and $\theta_{4}$ following the same steps as in sections \ref{InPoRRP} and \ref{InPoRRR}.

\subsection{R+R+R+R}
As in the previous case, the null point $P_{1}$ is obtained as the translation of $P_{0}$ along $\bm{z}_{1}$. Thus, the points that remain to be found are $P_{2}$ and $P_{3}$. Now, as stated in section \ref{InPoRRR}, two spheres and one plane are required to calculated $P_{3}$. One of such spheres is defined as in (\ref{S1RRPR}). For the other, we first consider a triangle whose sides are the second and third links. This triangle is similar to the one defined in the previous case: the side lengths are $d_{2}$ and $d_{3}$, respectively, and the angle between both sides is $\alpha_{3}$. The law of cosines allows to compute easily the length $d$ of the third side as:
\begin{equation}
d^2 = d_{3}^{2}+d_{2}^{2}-2d_{2}d_{3}\cos(\alpha_{3}),
\end{equation}
where only the positive solution for $d$ is taken. Now, the  inner representation of the other sphere can be defined:
\begin{equation}
s_{2} = P_{1} - \dfrac{1}{2}d^{2}e_\infty.
\end{equation}
On the other hand, the outer representation of a plane, denoted by $\Pi_{1}$, is computed as in (\ref{S1}). Clearly, $P_{3}$ can be extracted from the intersection bivector defined by these three geometric entities.

Once $P_{3}$ has been obtained, two new spheres should be defined to calculate $P_{2}$. Their inner representations are:
\begin{equation}
\begin{split}
s_{3} &= P_{1} - \dfrac{1}{2}d_{2}^{2}e_\infty,\\
s_{4} &= P_{3} - \dfrac{1}{2}d_{3}^{2}e_\infty.
\end{split}
\end{equation}
Again, the intersection of these two new spheres with the plane represented by $\Pi_{1}$ is computed as:
\begin{equation}
B = s_{3}\vee s_{4}\vee\pi_{1},
\end{equation}
where the null point $P_{2}$ is extracted from $B$. Finally, the joint variables $\theta_{1},\theta_{2},\theta_{3}$ and $\theta_{4}$ are found in a similar way as in \ref{InPoRRR} by means of $P_{0},P_{1},P_{2},P_{3}$ and $P_{w}$ (with all their possible combinations).

\section{Conclusions}\label{Conc.}
This paper proposes a geometric strategy based on conformal geometric algebra for solving the inverse kinematics of serial robotic manipulators with 6 and 7 DoF and a spherical wrist. For manipulators of this kind, the inverse kinematics can be decoupled into two subproblems: the position and the orientation problems. The proposed approach solves the first subproblem by developing particular geometric strategies for each combination of prismatic and revolute joints that describe the position part of the robot. On the other hand, the second subproblem is solved by splitting the rotor that defines the target orientation of the end-effector into exactly three rotors, where each one of them depends on just one joint variable.

For non-redundant serial robots, this approach gives the entire set of solutions, while for redundant robots with 7 DoF, all the solutions are obtained as a one-parameter family of particular solutions where the parameter is one of the joint variables (the redundant joint). Therefore, we can consider the approach introduced here as a closed-form method. Moreover, since offsets between consecutive joints in the position part of the robot are also treated, we are solving the problem for simple and complex robot geometries. As reviewed in sections \ref{Intro} and \ref{RW}, closed-form methods are the most suitable for solving the inverse kinematics of serial robots. Since the most existing contributions of this kind are based either in the use of complex geometric formulations or matrix manipulations, the approach presented here becomes an elegant, efficient and intuitive alternative.

\section*{References}

\bibliography{MiBibliografia}

\end{document}